%% file: main.tex
\documentclass[10pt,twocolumn,letterpaper]{article}
\usepackage[letterpaper,textwidth=6.875in,textheight=8.875in,top=1in,columnsep=0.3125in]{geometry}
\usepackage{times,xspace,graphicx,amsmath,amssymb,booktabs}
\usepackage[dvipsnames]{xcolor}
\usepackage[numbers,sort&compress]{natbib}
\usepackage[format=plain,labelformat=simple,labelsep=period,font=small,compatibility=false]{caption}
\usepackage[font=footnotesize,skip=3pt,subrefformat=parens]{subcaption}
\usepackage[shortlabels,inline]{enumitem}
\setlist[itemize]{noitemsep,leftmargin=*,topsep=0em}
\setlist[enumerate]{noitemsep,leftmargin=*,topsep=0em}
\usepackage{titlesec}
\usepackage{balance}
\newcommand{\eg}{\emph{e.g.}\xspace}

\usepackage[hyphens]{url}
\input{preamble}
\definecolor{linkblue}{rgb}{0.21,0.49,0.74}
\usepackage[breaklinks,colorlinks,allcolors=linkblue]{hyperref}
\usepackage[capitalize]{cleveref}
\crefname{section}{Sec.}{Secs.}
\Crefname{section}{Section}{Sections}
\crefname{table}{Tab.}{Tabs.}
\Crefname{table}{Table}{Tables}
\hypersetup{pdftitle={State-Message Interaction Adapter for Cross-Domain Few-Shot Segmentation},pdfauthor={Hang Xing, Yan Xia, Yongjiang Liu, Siqi Zhang, Xueming Ding},pdfsubject={Preprint},pdfkeywords={few-shot segmentation, parameter-efficient adaptation, DINOv3}}
\titleformat{\section}{\large\bfseries}{\thesection.}{0.25em}{}
\titleformat{\subsection}{\fontsize{11}{13}\selectfont\bfseries}{\thesubsection.}{0.25em}{}
\titleformat{\subsubsection}{\normalsize\bfseries}{\thesubsubsection.}{0.25em}{}
\titlespacing*{\section}{0pt}{10pt plus 2pt minus 2pt}{7pt}
\titlespacing*{\subsection}{0pt}{8pt plus 2pt minus 2pt}{5pt}
\titlespacing*{\subsubsection}{0pt}{6pt plus 2pt minus 2pt}{3pt}
\renewenvironment{abstract}{\centerline{\large\bfseries Abstract}\vspace{10pt}\noindent\itshape}{\par\vspace{10pt}}
\renewcommand{\topfraction}{0.99}
\renewcommand{\bottomfraction}{0.99}
\renewcommand{\textfraction}{0.01}
\renewcommand{\floatpagefraction}{0.99}
\renewcommand{\dbltopfraction}{0.99}
\renewcommand{\dblfloatpagefraction}{0.99}
\AtBeginDocument{\setlength{\abovedisplayskip}{4.5pt plus 1.5pt minus 1.5pt}\setlength{\belowdisplayskip}{4.5pt plus 1.5pt minus 1.5pt}}
\title{\bfseries State--Message Interaction Adapter for\\ Cross-Domain Few-Shot Segmentation}
\author{\begin{minipage}{\dimexpr\textwidth-2\tabcolsep\relax}\centering
\normalsize Hang Xing\textsuperscript{1,2}\quad Yan Xia\textsuperscript{1,2}\quad Yongjiang Liu\textsuperscript{1,2}\quad Siqi Zhang\textsuperscript{1,2}\quad Xueming Ding\textsuperscript{3,*}\\[7pt]
\small \textsuperscript{1}Wuxi Key Laboratory of Photovoltaic Equipment Advanced Manufacturing Technology, Wuxi, China\\[2pt]
\textsuperscript{2}Wuxi Autowell Technology Co., Ltd., Wuxi, China\\[2pt]
\textsuperscript{3}Department of Control Science and Engineering, School of Optical-Electrical and Computer Engineering,\\
University of Shanghai for Science and Technology, Shanghai, China\\[5pt]
\textsuperscript{*}Corresponding author.\quad Email (Hang Xing): \href{mailto:13605638511@163.com}{\texttt{13605638511@163.com}}
\end{minipage}}
\date{}
\begin{document}
\raggedbottom
\maketitle
\input{sec/0_abstract}
\input{sec/1_intro}
\input{sec/2_related}
\input{sec/3_method}
\input{sec/4_experiments}
\input{sec/5_conclusion}
{\small
\bibliographystyle{ieeenat_fullname}
\bibliography{main}}
\clearpage
\raggedbottom
\twocolumn[\begin{center}\Large\bfseries Appendix\end{center}]
\renewcommand{\topfraction}{0.95}
\renewcommand{\bottomfraction}{0.85}
\renewcommand{\textfraction}{0.05}
\renewcommand{\floatpagefraction}{0.90}
\renewcommand{\dbltopfraction}{0.92}
\renewcommand{\dblfloatpagefraction}{0.90}
\setcounter{topnumber}{3}
\renewcommand{\theHequation}{S\arabic{equation}}
\input{sec/X_suppl}
\end{document}

%% file: preamble.tex
\ifdefined\pdfsuppressptexinfo\pdfsuppressptexinfo=15\fi
\usepackage{microtype}
\usepackage{mathtools}
\usepackage{array}
\usepackage{colortbl}
\usepackage{pifont}

\definecolor{tblhead}{RGB}{241, 245, 249}
\definecolor{tblsub}{RGB}{248, 250, 252}
\definecolor{oursbg}{RGB}{228, 241, 252}
\definecolor{ourslight}{RGB}{242, 248, 254}
\definecolor{stdgray}{RGB}{70, 80, 95}
\definecolor{venuegray}{RGB}{105, 115, 130}
\definecolor{gaingreen}{RGB}{16, 124, 65}
\definecolor{markgreen}{RGB}{22, 135, 75}
\definecolor{markgray}{RGB}{165, 172, 182}

\newcommand{\std}[1]{{\footnotesize\textcolor{stdgray}{\,$\pm$\,#1}}}
\newcommand{\bstd}[1]{{\footnotesize\textcolor{stdgray}{\,$\pm$\,#1}}}
\newcommand{\venue}[1]{{\scriptsize\textcolor{venuegray}{[#1]}}}
\newcommand{\gain}[1]{{\scriptsize\textcolor{gaingreen}{\textbf{(+#1)}}}}
\newcommand{\cmark}{{\textcolor{markgreen}{\ding{51}}}}
\newcommand{\xmark}{{\textcolor{markgray}{\ding{55}}}}
\newcommand{\secbest}[1]{\underline{#1}}

\newcommand{\R}{\mathbb{R}}
\newcommand{\diag}{\operatorname{diag}}
\newcommand{\vecoff}{\operatorname{vec}_{\mathrm{off}}}

%% file: sec/0_abstract.tex
\begin{abstract}
Cross-domain few-shot segmentation adapts natural-image pretrained encoders to medical or industrial domains from a few pixel-annotated support masks. Within frozen DINOv3, intermediate per-head attention messages already carry target--background distinctions that baseline final-layer similarities and predictions do not consistently reflect. We propose the State--Message Interaction Adapter (SMIA) to learn how per-head messages should update each receiving token representation while keeping pretrained attention and FFN parameters frozen. Inserted right after per-head aggregation, SMIA conditions the transformation of each head's aggregated output (\emph{message}) on each patch's pre-attention feature (\emph{state}), allowing similar contextual messages to produce distinct or opposing updates depending on the receiving position. Because scarce support masks cover only limited state--message combinations, a shared low-rank bilinear map confines the learned interactions to low-dimensional state, message, and output subspaces. Added before each adapted block's frozen FFN, these updates yield adapted features that are refined by frozen FFNs and deeper blocks for linear readout. With only 0.51M trainable parameters and single-pass, support-free inference, SMIA achieves six-dataset mean mIoUs of 77.40\% at 5 shots and 80.99\% at 10 shots across 20 medical and industrial tasks, surpassing the strongest prior methods by +3.61 and +3.93 percentage points. In controlled 10-shot comparisons across three benchmarks on the same DINOv3 backbone, SMIA also outperforms parameter-matched AdaptFormer, LoRA, and ParaX with a better precision--recall balance.
\end{abstract}

%% file: sec/1_intro.tex
\section{Introduction}
\label{sec:intro}

\begin{figure}[!tbp]
  \centering
  \includegraphics[width=0.94\linewidth]{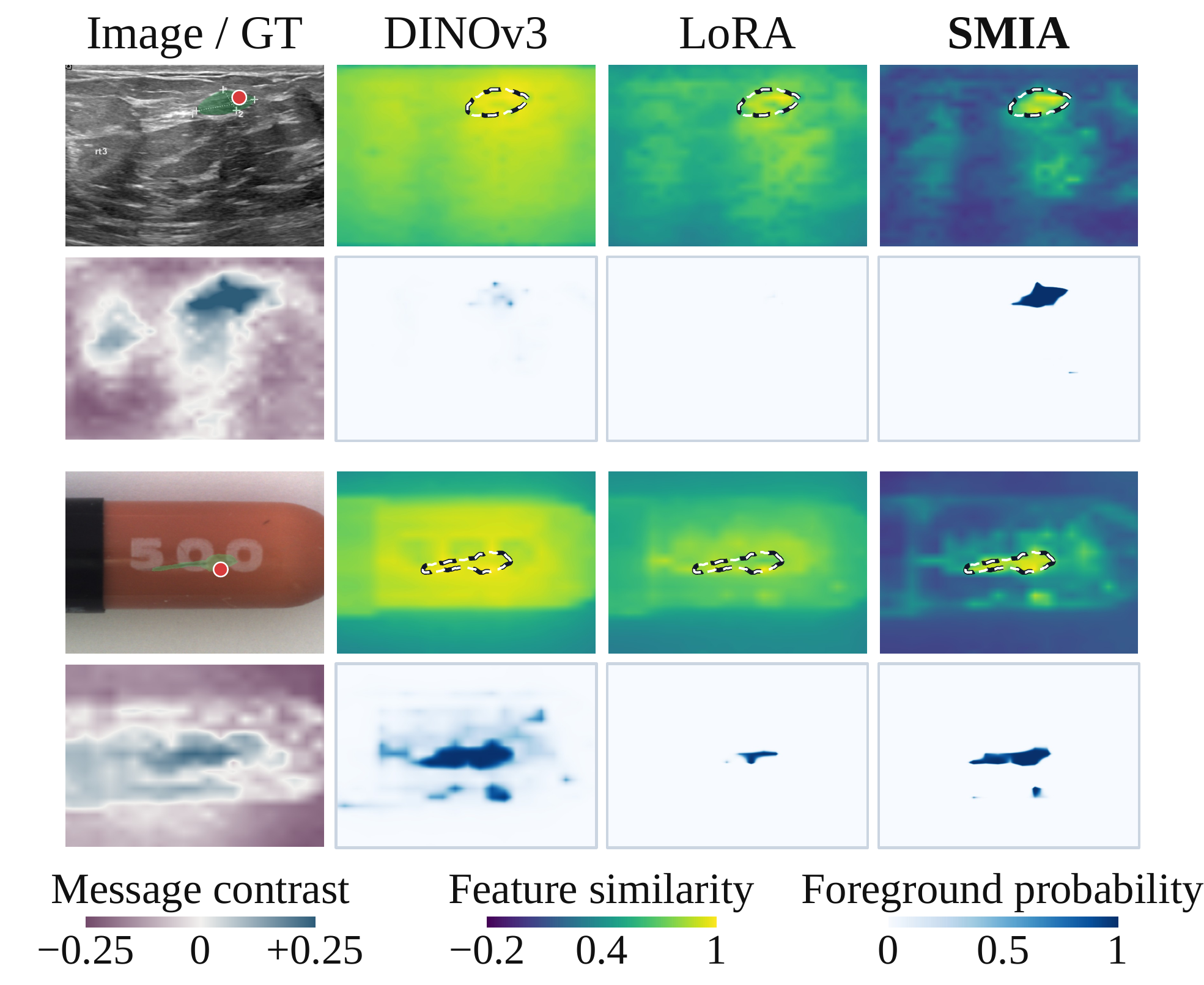}
  \caption{\textbf{Intermediate attention messages vs.\ downstream segmentation responses.} Left column: query image with ground-truth mask and marked foreground patch (top), plus single-head message contrast from frozen DINOv3 Block~9 (bottom; Head~2 on BUSI, Head~5 on MVTec Capsule), where teal/plum indicates closer cosine proximity to foreground/adjacent-background references. Right three columns: final-layer cosine similarity to the marked patch with dashed white ground-truth contours (top) and foreground probability from linear readouts fit on the same 10 supports (bottom; diagnostic 4-block SMIA and rank-8 LoRA setup in Supplementary Sec.~C.1). Only SMIA recovers clean, localized boundaries.}
  \label{fig:motivation}
  \vspace{-4pt}
\end{figure}

Acquiring pixel-level masks is costly in both clinical imaging and industrial inspection. Delineating subtle lesions requires specialist time, while newly introduced manufactured parts rarely come with more than a few defective examples. Cross-domain few-shot segmentation (CD-FSS) addresses this setting by transferring natural-image pretrained encoders to domains such as ultrasound, endoscopy, dermoscopy, radiography, and textured industrial surfaces from only a handful of annotated support masks~\cite{lei2022cdfss,ma2026hera,zhang2025sofs}.

A few annotated masks rarely cover enough appearance variation to train a dense encoder from scratch. Self-supervised vision foundation models (VFMs) such as DINOv3~\cite{simeoni2026dinov3} shift the few-shot burden from representation learning to adapting and reading out pretrained features, providing dense patch embeddings and multi-head context aggregation before target adaptation. Yet medical lesions and fine surface cracks often share intensity, texture, or structural continuity with surrounding healthy tissue or intact material. Features that group patches by object part or material identity need not delineate the target boundaries required downstream, and few-shot adaptation must learn this distinction from support masks covering only limited appearances and contexts.

Recent VFM-based segmenters either match frozen features against support exemplars without training (\eg, INSID3~\cite{cuttano2026insid3}) or train multi-scale decoders on frozen layer outputs (\eg, DINO-AugSeg~\cite{xu2026dinoaugseg}). To examine how intermediate pretrained cues carry over to downstream segmentation, \cref{fig:motivation} compares a frozen DINOv3 probe, LoRA~\cite{hu2022lora}, and our adapter, each paired with a linear readout head fit on the same 10-shot support set. On BUSI ultrasound, a single frozen Block-9 head already separates the lesion from surrounding stroma, yet both baselines produce diffuse final-layer similarities and miss nearly the entire lesion. On MVTec Capsule, the frozen head responds to both the crack and nearby printed digits; the frozen probe over-segments the surrounding shell, whereas LoRA retains only a small fragment of the crack.

These examples show that target-relevant structure in intermediate messages is not consistently reflected in final predictions, motivating adaptation of how intermediate messages contribute to token updates. Within an attention block, a head's \emph{aggregated message} summarizes contextual cues pooled for a location, whereas the pre-attention \emph{local state} carries the receiving patch's own content and prior context. This distinction matters when adjacent target and background patches receive strongly aligned messages yet differ in their pre-attention states, as illustrated by the boundary pairs in \cref{sec:mechanism}.

We propose the State--Message Interaction Adapter (SMIA), which keeps pretrained attention and FFN parameters frozen and inserts a conditional branch right \emph{after} per-head aggregation. Whereas LoRA applies token-invariant low-rank increments to selected projections, SMIA parameterizes the added transformation of each head's message as a linear function of the receiving patch's local state. Similar contextual messages can therefore yield distinct or opposing updates across positions, directing target supervision toward how pooled contextual cues modify each receiving representation.

Parameterizing this conditional relation freely in ambient dimensions would require $O(d^2 d_h)$ weights per head, whereas scarce support masks cover only limited combinations of content and context. SMIA instead restricts the interaction to low-dimensional state, message, and output subspaces through a structured bilinear map shared across patches and images within a task. We add these state-guided updates before each adapted block's frozen feed-forward network (FFN); frozen FFNs and deeper blocks then refine the updated features for support-free inference with a single linear layer.

Across 20 tasks spanning five medical benchmarks and all 15 MVTec AD categories, SMIA sets a new state of the art at both 5 and 10 shots while maintaining over $82\%$ industrial precision alongside substantially higher defect recall (\cref{sec:main_results}). Our main contributions are:
\begin{itemize}
  \item We propose SMIA, an in-block post-aggregation adapter that keeps pretrained attention and FFN parameters frozen and conditions per-head message updates on each patch's local state within shared low-rank subspaces, with 0.51M trainable parameters, $O(N_p)$ overhead, and single-pass, support-free inference.
  \item We validate this state-dependent mechanism through forward-pass boundary diagnostics and controlled interventions: SMIA maps strongly aligned boundary messages to opposing updates, and norm-matched counterfactual state swaps under fixed messages flip local foreground--background predictions.
  \item Across 20 medical and industrial tasks, SMIA improves six-dataset mean mIoU over the strongest prior methods by $+3.61$ and $+3.93$ percentage points at 5 and 10 shots, ranking first on 16/20 and 18/20 tasks. In controlled 10-shot comparisons across three benchmarks on the same DINOv3 backbone, SMIA achieves higher mIoU than frozen readout and parameter-matched AdaptFormer, LoRA, and ParaX, with a better precision--recall balance than the three PEFT baselines.
\end{itemize}

%% file: sec/2_related.tex
\section{Related Work}
\label{sec:related}

\paragraph{Cross-domain few-shot segmentation.}
CD-FSS transfers natural-image pretrained segmenters to target domains with large distribution shifts using a few support masks~\cite{lei2022cdfss}. One line of work redesigns cross-image support--query matching, such as self-disentangling and recomposing multi-layer features in SDRC~\cite{tong2025sdrc}. Other methods adapt representations to the target domain: HERA combines layer selection, regularized target tuning, and pixelwise calibration~\cite{ma2026hera}, while MPA uses multi-view progressive adaptation~\cite{nie2026mpa}. Related industrial methods target small surface defects by suppressing background clutter during matching in SOFS~\cite{zhang2025sofs} or adapting SAM hierarchically in DefectSAM~\cite{yan2025defectsam}. At test time, however, these approaches still depend on cross-image exemplar matching, prototype calibration, or multi-scale decoders. By contrast, SMIA learns in-block residual updates from the support masks for single-pass, support-free segmentation.

\begin{figure*}[!tbp]
  \centering
  \includegraphics[width=0.88\textwidth]{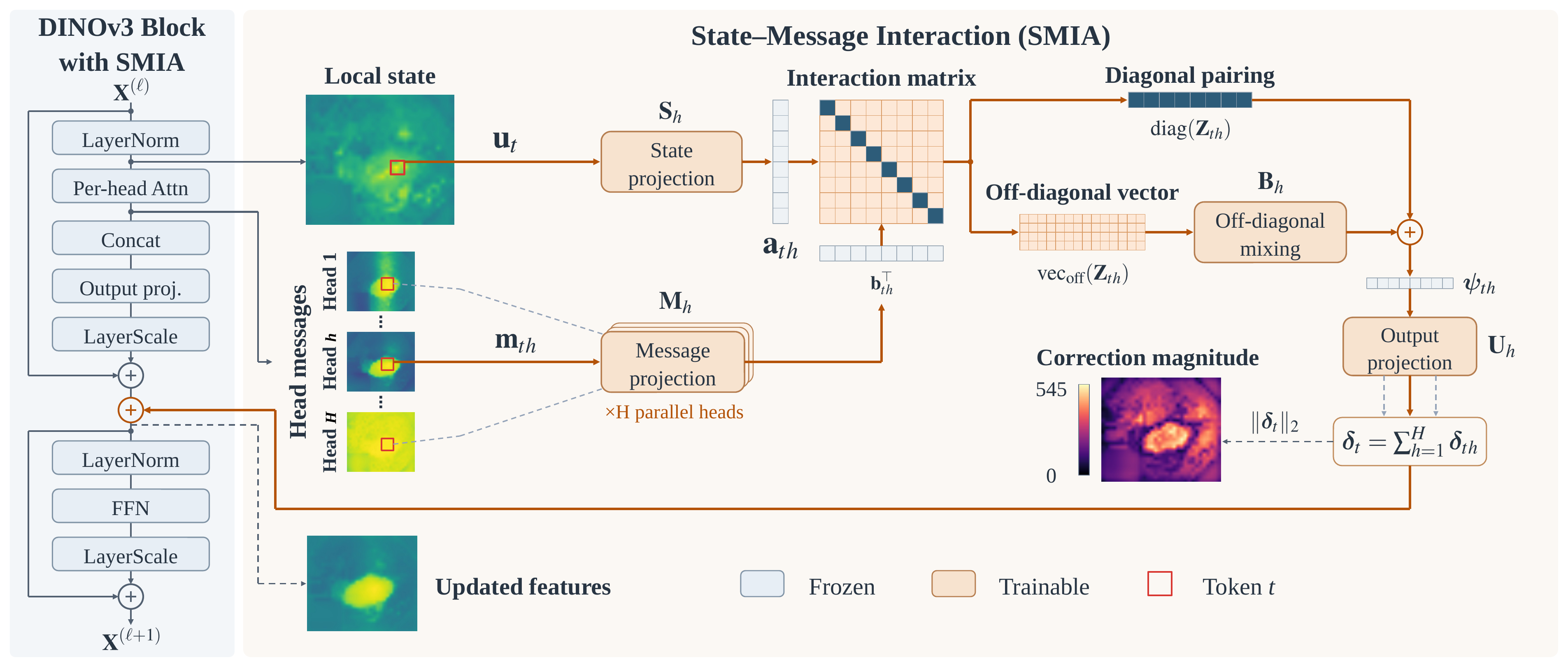}
  \caption{\textbf{In-block architecture and state--message interaction of SMIA.} \emph{Left:} Frozen DINOv3 block with SMIA patch correction injected before the FFN sublayer. \emph{Right:} Conditional update path for patch~$t$, illustrated on a Kvasir-SEG query at Block~12 of 4-block SMIA: heatmaps show cosine similarity to reference patch~$t$ (shared $[-1, 1]$ scale; brighter is higher) for local state $\mathbf{u}_t$, per-head messages $\mathbf{m}_{th}$, and pre-FFN updated features combining the original attention residual with $\boldsymbol{\delta}_t$, while correction magnitude plots $\|\boldsymbol{\delta}_t\|_2$ ($[0, 545]$).}
  \label{fig:architecture}
  \vspace{-4pt}
\end{figure*}

\paragraph{Segmentation with vision foundation models.}
Self-supervised VFMs such as DINOv3~\cite{simeoni2026dinov3} provide spatially coherent patch features that markedly improve few-shot segmentation. Training-free matchers infer masks directly from frozen feature geometry: INSID3 combines positional debiasing, intra-query clustering, and exemplar matching~\cite{cuttano2026insid3}, whereas RSRM re-fuses multi-layer features for region matching~\cite{sun2026rsrm}. Supervised decoder methods keep the encoder frozen and train backend readouts; DINO-AugSeg pairs Haar wavelet feature augmentation with a cross-guided U-Net on frozen DINOv3~\cite{xu2026dinoaugseg}, and DistillFSS distills support--query fusion into a standalone network~\cite{demarinis2025distillfss}. In both families, support masks inform prediction only after the frozen encoder finishes token processing. SMIA instead adapts token updates inside each Transformer block right after per-head aggregation, allowing frozen FFNs and deeper blocks to refine task-specific features progressively.

\paragraph{Parameter-efficient adaptation for vision.}
Parameter-efficient fine-tuning (PEFT) adapts frozen encoders through compact trainable modules, such as low-rank projection updates in LoRA~\cite{hu2022lora} or parallel FFN bottlenecks in AdaptFormer~\cite{chen2022adaptformer}. For dense prediction and domain generalization, Rein couples learnable token banks to patch features~\cite{wei2024rein}, while ParaX routes shared expert weights at the image level~\cite{lou2026parax}. Recently, Gated Attention~\cite{qiu2025gated} modulates per-head attention outputs with query-dependent sigmoid gates during language model pretraining. In frozen-backbone adaptation, however, LoRA applies token-invariant low-rank increments to selected projections, and AdaptFormer, Rein, and ParaX adapt block-level token features without conditioning per-head message updates on pre-attention local state. SMIA keeps pretrained attention and FFN parameters frozen, using each patch's local state right after per-head aggregation to weight shared low-rank message matrices for signed directional updates.

%% file: sec/3_method.tex
\section{State--Message Interaction Adapter}
\label{sec:method}

SMIA keeps all pretrained attention and FFN parameters frozen and adds a lightweight post-aggregation branch inside each adapted Transformer block (\cref{fig:architecture}). Rather than tuning attention parameters, this branch conditions each head's residual update on the patch's pre-attention local state within learned low-rank subspaces. Below we present the state-guided post-aggregation update (\cref{sec:formulation}), its low-rank bilinear parameterization (\cref{sec:factorization}), the resulting state-dependent basis combination (\cref{sec:operator}), and few-shot training and inference (\cref{sec:optimization}).

\subsection{State-Guided Post-Aggregation Update}
\label{sec:formulation}

Given a target support set $\mathcal{S} = \{(I_n, Y_n)\}_{n=1}^N$ of $N$ pixel-annotated images, SMIA adapts intermediate patch representations inside the frozen pretrained encoder. Although SMIA can be combined with heavier segmentation decoders, we pair the adapted encoder with a single linear layer trained on $\mathcal{S}$ so representation quality is evaluated directly without extra decoder capacity. All remaining images in each task inventory serve as test queries (\cref{sec:protocol}).

Consider one adapted Transformer block (\cref{fig:architecture}) with $H$ attention heads and $N_p$ spatial patch tokens indexed by $t \in \{1, \dots, N_p\}$ and $h \in \{1, \dots, H\}$. Let $\mathbf{u}_t \in \R^d$ denote the pre-attention LayerNorm output of patch $t$, which we refer to as its \emph{local state}. For head $h$, let $\mathbf{m}_{th} \in \R^{d_h}$, with per-head dimension $d_h = d / H$, denote the attention-weighted value sum before head concatenation and output projection; we call $\mathbf{m}_{th}$ the head's \emph{aggregated message}. A fixed linear projection applies the same position-agnostic map to any incoming message. By contrast, SMIA maps $\mathbf{m}_{th}$ to a residual update $\boldsymbol{\delta}_{th} \in \R^d$ through a linear transformation determined by the receiving state $\mathbf{u}_t$:
\begin{equation}
  \boldsymbol{\delta}_{th} = \mathbf{K}_h(\mathbf{u}_t)\mathbf{m}_{th}, \qquad \mathbf{K}_h(\mathbf{u}_t) \in \R^{d \times d_h},
  \label{eq:cond_map}
\end{equation}
where $\mathbf{K}_h(\mathbf{u}_t)$ is the local linear map induced at state $\mathbf{u}_t$. Adapter weights are distinct across blocks and heads, yet shared across all patches and images within a task so every labeled support pixel constrains the same conditional update rule.

\subsection{Low-Rank Bilinear Parameterization}
\label{sec:factorization}

We model the linear dependence of $\mathbf{K}_h(\mathbf{u}_t)$ on $\mathbf{u}_t$ through bilinear interaction~\cite{pan2020xlinear}. Parameterizing an unconstrained mapping in ambient dimensions would require $d^2 d_h$ weights per head, which amounts to $9.44\mathrm{M}$ parameters per head on ViT-S+/16 and is impractical for few-shot adaptation. SMIA instead projects both the local state and the aggregated message onto $r$ unit-norm directions with $r \ll \min(d, d_h)$:
\begin{equation}
  \mathbf{a}_{th} = \mathbf{S}_h \mathbf{u}_t, \qquad \mathbf{b}_{th} = \mathbf{M}_h \mathbf{m}_{th}, \qquad \mathbf{a}_{th}, \mathbf{b}_{th} \in \R^r,
  \label{eq:proj}
\end{equation}
using learnable projection matrices $\mathbf{S}_h \in \R^{r \times d}$ and $\mathbf{M}_h \in \R^{r \times d_h}$.

Across Transformer depths and heads, local states $\mathbf{u}_t$ and messages $\mathbf{m}_{th}$ vary substantially in coordinate variance. Without normalization, high-variance factor pairs $a_{thi}b_{thj}$ can dominate the initial gradients on $\mathbf{U}_h$ and unbalance subsequent bilinear updates. To equalize second-order product scales before mixing, SMIA divides the pairwise outer product elementwise by a fixed positive RMS calibration matrix $\mathbf{C}_h \in \R^{r \times r}$ with entries $c_{hij} > 0$, estimated once on the unaugmented support set before training (Supplementary Sec.~A.2). Combining coordinate-wise pairing~\cite{kim2017hadamard}, off-diagonal mixing $\mathbf{B}_h \in \R^{r \times [r(r-1)]}$, and output projection $\mathbf{U}_h \in \R^{d \times r}$, the per-head update takes the form:
\begin{equation}
  \begin{aligned}
    \mathbf{Z}_{th} &= (\mathbf{a}_{th}\mathbf{b}_{th}^\top) \oslash \mathbf{C}_h, \\
    \boldsymbol{\psi}_{th} &= \diag(\mathbf{Z}_{th}) + \mathbf{B}_h \vecoff(\mathbf{Z}_{th}), \\
    \boldsymbol{\delta}_{th} &= \mathbf{U}_h \boldsymbol{\psi}_{th},
  \end{aligned}
  \label{eq:interaction}
\end{equation}
where $\oslash$ denotes elementwise division, $\diag(\cdot) \in \R^r$ extracts diagonal pairs, and $\vecoff(\cdot) \in \R^{r(r-1)}$ flattens off-diagonal entries. The diagonal terms $\diag(\mathbf{Z}_{th})$ modulate each message coordinate directly by its paired state coordinate, whereas $\mathbf{B}_h$ captures cross-coordinate coupling without widening the bottleneck~$r$. Together, Eqs.~\eqref{eq:proj}--\eqref{eq:interaction} form a structured rank-$(r,r,r)$ Tucker factorization~\cite{benyounes2017mutan} that reduces per-head parameters from $d^2 d_h$ to $r(2d + d_h + r(r-1))$.

\subsection{State-Dependent Basis Combination and Integration}
\label{sec:operator}

Rearranging \cref{eq:interaction} with respect to the projected state coordinates $a_{thi} = (\mathbf{S}_h \mathbf{u}_t)_i$ expresses $\mathbf{K}_h(\mathbf{u}_t)$ as a signed linear combination of $r$ shared basis matrices $\{\mathbf{L}_{hi}\}_{i=1}^r$:
\begin{equation}
  \mathbf{K}_h(\mathbf{u}_t) = \sum_{i=1}^r (\mathbf{S}_h \mathbf{u}_t)_i \, \mathbf{L}_{hi},
  \label{eq:operator_bank}
\end{equation}
where each $\mathbf{L}_{hi} \in \R^{d \times d_h}$ is determined by $\mathbf{U}_h, \mathbf{B}_h, \mathbf{C}_h$, and $\mathbf{M}_h$.
This decomposition clarifies how shared message basis matrices produce state-dependent updates from a compact parameter set. During training, all labeled support pixels jointly fit the same $r$ basis matrices $\{\mathbf{L}_{hi}\}_{i=1}^r$. At test time, $\{\mathbf{L}_{hi}\}_{i=1}^r$ remain fixed while local coordinates $(\mathbf{S}_h \mathbf{u}_t)_i$ assemble a patch-specific linear map $\mathbf{K}_h(\mathbf{u}_t)$ of rank at most $r$, without modifying pretrained attention parameters. Because the coefficients $(\mathbf{S}_h \mathbf{u}_t)_i$ carry signs, patches with distinct local states can map similar attention messages $\mathbf{m}_{th}$ to distinct or opposing residual updates.

Summing the head updates $\boldsymbol{\delta}_t = \sum_{h=1}^H \boldsymbol{\delta}_{th}$ into a patch matrix $\boldsymbol{\Delta}_\ell(\mathbf{X}^{(\ell)})$ and zero-padding class and register tokens via $\mathcal{P}_{\mathrm{patch}}$ gives the adapted block update (\cref{fig:architecture}, left):
\begin{equation}
  \mathbf{X}^{(\ell+1)} = \mathcal{R}_0^\ell\!\left(\mathcal{A}_0^\ell(\mathbf{X}^{(\ell)}) + \mathcal{P}_{\mathrm{patch}} \boldsymbol{\Delta}_\ell(\mathbf{X}^{(\ell)})\right),
  \label{eq:block}
\end{equation}
which places the patch correction between the frozen attention residual $\mathcal{A}_0^\ell$ and the frozen FFN residual $\mathcal{R}_0^\ell$ so downstream FFNs and deeper blocks remain intact.

\begin{table*}[!tbp]
  \caption{\textbf{Quantitative comparison across 20 medical and industrial tasks under 5-shot and 10-shot support budgets.} Values report native-resolution mIoU (\%, mean $\pm$ SD across five nested random support draws). \emph{MVTec AD} averages all 15 industrial categories equally per draw, and \emph{Overall} weights the five medical datasets and MVTec AD equally (green parentheses: gains over the strongest prior method computed before rounding). Best and second-best scores per budget are \textbf{bolded} and \secbest{underlined}.}
  \label{tab:external}
  \centering
  {\setlength{\aboverulesep}{0pt}\setlength{\belowrulesep}{0pt}\renewcommand{\arraystretch}{1.18}\setlength{\tabcolsep}{3.4pt}%
  \resizebox{\textwidth}{!}{%
  \begin{tabular}{llccccccc}
    \toprule
    \rowcolor{tblhead}
    & & \multicolumn{5}{c}{\textbf{Medical Imaging Benchmarks (5 Tasks)}} & \textbf{Industrial (15)} & \textbf{6-Dataset Summary} \\
    \cmidrule(lr){3-7} \cmidrule(lr){8-8} \cmidrule(lr){9-9}
    \rowcolor{tblhead}
    \textbf{Method} & \textbf{Backbone} & \textbf{BUSI} & \textbf{Kvasir-SEG} & \textbf{CVC-ClinicDB} & \textbf{ISIC 2018} & \textbf{Chest X-ray} & \textbf{MVTec AD} & \textbf{Overall mIoU} \\
    \midrule
    \rowcolor{tblsub}
    \multicolumn{9}{l}{\textit{\textbf{5-Shot Support Budget ($N = 5$)}}} \\
    \midrule
    INSID3~\cite{cuttano2026insid3}~\venue{CVPR'26}     & DINOv3 ViT-L/16  & 40.01\std{6.02} & 81.07\std{1.03} & 63.08\std{8.80} & \secbest{76.13}\std{1.08} & 85.65\std{0.27} & 56.75\std{1.76} & 67.11\std{1.90} \\
    RSRM~\cite{sun2026rsrm}~\venue{ECCV'26}             & DINOv3 ViT-B/16  & 61.49\std{1.63} & \textbf{81.85}\std{0.70} & 73.39\std{4.95} & 75.86\std{1.16} & 90.86\std{1.64} & 52.62\std{0.95} & 72.68\std{0.79} \\
    DINO-AugSeg~\cite{xu2026dinoaugseg}~\venue{MLST'26} & DINOv3 ConvNeXt-L & 56.83\std{6.41} & 76.54\std{3.75} & 74.35\std{5.21} & 71.15\std{6.20} & \secbest{92.78}\std{0.90} & \secbest{66.40}\std{1.72} & 73.01\std{1.43} \\
    DefectSAM~\cite{yan2025defectsam}~\venue{TNNLS'25}  & SAM ViT-B        & 55.87\std{6.74} & 60.88\std{4.14} & 58.53\std{3.71} & 71.80\std{2.01} & 88.30\std{5.73} & 62.16\std{0.73} & 66.25\std{1.57} \\
    HERA~\cite{ma2026hera}~\venue{CVPR'26}              & DINOv3 ViT-L/16  & \secbest{61.91}\std{5.66} & \secbest{81.62}\std{1.76} & \textbf{76.52}\std{7.34} & 75.43\std{2.32} & 91.81\std{0.89} & 55.45\std{2.26} & \secbest{73.79}\std{1.20} \\
    MPA~\cite{nie2026mpa}~\venue{CVPR'26}               & ResNet-50        & 57.17\std{5.73} & 70.22\std{3.17} & 68.99\std{2.03} & 73.35\std{2.00} & 92.58\std{0.61} & 65.14\std{0.48} & 71.24\std{0.87} \\
    \rowcolor{oursbg}
    \textbf{SMIA (Ours)}                                & \textbf{DINOv3 ViT-S+/16} & \textbf{62.05}\bstd{4.51} & 80.28\std{2.10} & \secbest{76.34}\std{6.22} & \textbf{79.98}\bstd{1.22} & \textbf{94.38}\bstd{0.60} & \textbf{71.38}\bstd{1.70} & \textbf{77.40}\bstd{0.28}~\gain{3.61} \\
    \midrule
    \rowcolor{tblsub}
    \multicolumn{9}{l}{\textit{\textbf{10-Shot Support Budget ($N = 10$)}}} \\
    \midrule
    INSID3~\cite{cuttano2026insid3}~\venue{CVPR'26}     & DINOv3 ViT-L/16  & 37.21\std{4.47} & 81.72\std{0.56} & 65.68\std{4.34} & \secbest{76.80}\std{0.22} & 85.64\std{0.08} & 57.98\std{1.41} & 67.51\std{1.07} \\
    RSRM~\cite{sun2026rsrm}~\venue{ECCV'26}             & DINOv3 ViT-B/16  & 64.22\std{1.16} & \secbest{83.00}\std{1.18} & 74.24\std{3.55} & 75.92\std{0.55} & 91.45\std{0.58} & 52.40\std{0.80} & 73.54\std{0.82} \\
    DINO-AugSeg~\cite{xu2026dinoaugseg}~\venue{MLST'26} & DINOv3 ConvNeXt-L & \secbest{66.59}\std{2.05} & 81.10\std{1.61} & 77.52\std{2.05} & 72.21\std{5.73} & \secbest{93.37}\std{0.92} & \secbest{71.58}\std{1.51} & \secbest{77.06}\std{0.93} \\
    DefectSAM~\cite{yan2025defectsam}~\venue{TNNLS'25}  & SAM ViT-B        & 64.20\std{1.56} & 66.79\std{2.94} & 64.45\std{1.80} & 75.08\std{1.81} & 93.12\std{1.12} & 68.30\std{0.51} & 71.99\std{0.50} \\
    HERA~\cite{ma2026hera}~\venue{CVPR'26}              & DINOv3 ViT-L/16  & 65.34\std{2.75} & \textbf{83.45}\std{0.80} & \secbest{79.78}\std{6.47} & 76.35\std{0.49} & 93.10\std{0.50} & 58.30\std{1.52} & 76.05\std{1.11} \\
    MPA~\cite{nie2026mpa}~\venue{CVPR'26}               & ResNet-50        & 62.90\std{2.07} & 73.46\std{4.30} & 72.20\std{1.91} & 76.16\std{1.55} & 92.93\std{0.59} & 67.62\std{0.48} & 74.21\std{1.05} \\
    \rowcolor{oursbg}
    \textbf{SMIA (Ours)}                                & \textbf{DINOv3 ViT-S+/16} & \textbf{70.32}\bstd{1.34} & 82.21\std{3.38} & \textbf{81.42}\bstd{2.64} & \textbf{80.39}\bstd{1.64} & \textbf{94.63}\bstd{0.49} & \textbf{76.96}\bstd{1.11} & \textbf{80.99}\bstd{1.10}~\gain{3.93} \\
    \bottomrule
  \end{tabular}%
  }}
  \vspace{-4pt}
\end{table*}

\subsection{Few-Shot Training and Inference}
\label{sec:optimization}

During training, foreground patch logits from the linear head are bilinearly upsampled to the input canvas, unpadded, and resized to native image resolution. We minimize a class-balanced pixelwise BCE loss that averages within each present class of an image and weights present classes and sampled supports equally. At test time, the same restored native-resolution logits are thresholded at $0$, corresponding to foreground probability $p > 0.5$.

Our optimization schedule follows the gradient and scale structure of \cref{eq:interaction}. We initialize $\mathbf{U}_h = \mathbf{0}$ and $\mathbf{B}_h = \mathbf{0}$ while drawing unit-norm Gaussian rows for $\mathbf{S}_h$ and $\mathbf{M}_h$, ensuring that training starts from exact frozen DINOv3 representations. At $\mathbf{U}_h = \mathbf{0}$, step-1 adapter gradients are nonzero only on $\mathbf{U}_h$ and depend on the backpropagated readout error. Zero classifier weights would block initial gradients into the adapter branch, whereas random classifier weights inject noisy gradient directions on scarce supports. We therefore pre-fit the linear head on frozen unaugmented support features via convex $\ell_2$-regularized logistic regression at $\lambda = 10^{-3}$, providing task-aligned gradients from step~1.

We then drop the ridge penalty, jointly optimize all adapter and classifier parameters for 2,000 Adam steps at learning rate $10^{-4}$, and evaluate the final step-2,000 checkpoint directly (solver and sampling details appear in Supplementary Sec.~A.2). Unconstrained row-norm drift in $\mathbf{S}_h$ or $\mathbf{M}_h$ would introduce scale redundancy with $(\mathbf{B}_h, \mathbf{U}_h)$ relative to $\mathbf{C}_h$. After each Adam step, we project the raw displacement $\boldsymbol{\Delta}$ for each unit row $\mathbf{w}$ of $\mathbf{S}_h$ or $\mathbf{M}_h$ onto the tangent plane at $\mathbf{w}$, and then retract the updated row onto the unit sphere:
\begin{equation}
  \boldsymbol{\Delta}_\perp = \boldsymbol{\Delta} - (\boldsymbol{\Delta}^\top \mathbf{w})\mathbf{w}, \qquad \mathbf{w} \leftarrow \frac{\mathbf{w} + \boldsymbol{\Delta}_\perp}{\|\mathbf{w} + \boldsymbol{\Delta}_\perp\|_2},
  \label{eq:retract}
\end{equation}
which bounds row-norm drift under the fixed calibration scales $\mathbf{C}_h$.

On DINOv3 ViT-S+/16 ($L=12$, $d=384$, $H=6$, $d_h=64$) at $r=8$, each head introduces $7{,}104$ weights, comprising $3{,}072$ in $\mathbf{S}_h$, $512$ in $\mathbf{M}_h$, $448$ in $\mathbf{B}_h$, and $3{,}072$ in $\mathbf{U}_h$. Together with the $385$-parameter linear head, the model has 511,873 trainable parameters in total. Every adapter weight is applied once per patch token, adding $O(N_p)$ overhead of $0.51\mathrm{M}$ MACs per patch—or $0.52$ and $1.18$~GMACs on $512^2$ and $768^2$ canvases—independent of support size~$N$ for fast support-free inference (Supplementary Tab.~S3).

%% file: sec/4_experiments.tex
\section{Experiments and Analysis}
\label{sec:experiments}

We describe our 20-task evaluation protocol in \cref{sec:protocol}, compare SMIA against six recent methods in \cref{sec:main_results}, present controlled same-backbone ablations in \cref{sec:ablations}, and verify the state-guided update mechanism in \cref{sec:mechanism}.

\subsection{Experimental Setup and Unified Protocol}
\label{sec:protocol}

\paragraph{Tasks and support sampling.}
Our benchmark covers 20 segmentation tasks spanning five medical datasets and all 15 object and texture anomaly categories of MVTec AD~\cite{bergmann2019mvtec}. The medical benchmarks comprise BUSI breast ultrasound lesions~\cite{aldhabyani2020busi}, Kvasir-SEG~\cite{jha2020kvasir} and CVC-ClinicDB~\cite{bernal2015wmdova} colonoscopy polyps, ISIC 2018 Task~1 dermoscopy lesions~\cite{codella2019isic}, and Chest X-ray dual-lung segmentation pooling Shenzhen and Montgomery radiographs~\cite{jaeger2014shenzhen,candemir2014montgomery}. Main 20-task evaluations (\cref{tab:external,tab:ind_pr}) test $N=5$ and $N=10$ supports across five random draws, where each 5-shot set is strictly nested inside its 10-shot counterpart. All 20 tasks use image-level splits that treat all non-support images in the task inventory as test queries. Specifically, ISIC pools the official training, validation, and test splits while sampling supports strictly from the training split, and MVTec AD likewise pools all official splits. On BUSI and MVTec AD, which contain target-free normal scans, $N=5$ and $N=10$ enforce positive/negative quotas of $3/2$ and $6/4$ and count normal all-background masks toward budget~$N$ (Supplementary Sec.~A.1).

\begin{figure*}[!tbp]
  \centering
  \includegraphics[width=0.84\textwidth]{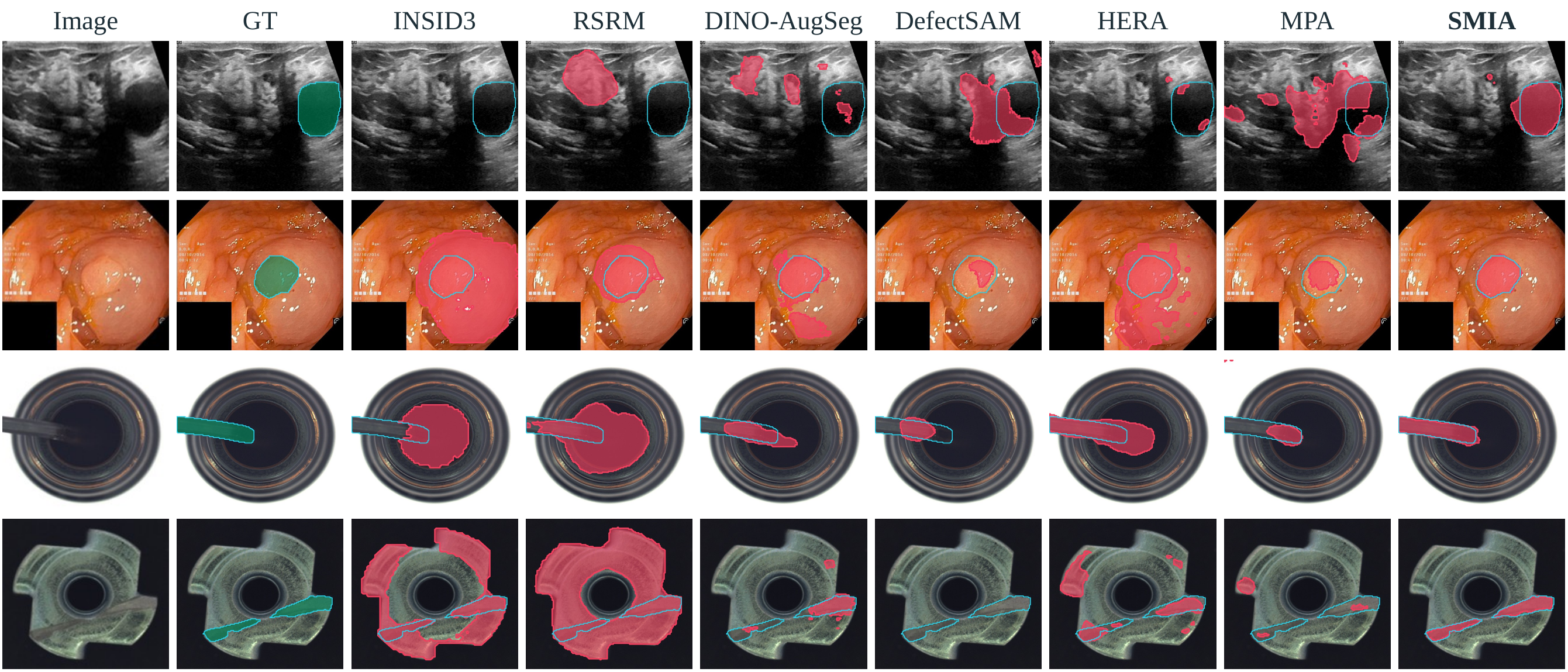}
  \caption{\textbf{Qualitative comparison across medical and industrial tasks ($N=10$, shared support draw per row).} Top to bottom: BUSI breast ultrasound, Kvasir-SEG polyp, MVTec Bottle contamination, and MVTec Metal Nut scratch (cyan contours: ground-truth boundaries; coral-red overlays: predicted foreground). SMIA delineates fine contours without over-segmenting surroundings.}
  \label{fig:qualitative}
  \vspace{-6pt}
\end{figure*}

\paragraph{Evaluation metrics and implementation.}
For every task, budget, and draw, all compared methods receive the exact same support set and are evaluated on the identical full query set at native image resolution. Exemplar-matching baselines (INSID3, RSRM, and HERA) construct reference templates from the positive supports while normal supports still count toward budget~$N$ (ground-truth preprocessing details appear in Supplementary Sec.~A.1). Within each draw, pixel confusion counts $\mathrm{TP}, \mathrm{FP}, \mathrm{FN}, \mathrm{TN}$ are accumulated across all native-resolution query pixels, and task mIoU averages global foreground and background IoUs. In \cref{tab:external}, MVTec AD averages all 15 industrial classes equally per draw, and the six-dataset \emph{Overall} mIoU weights MVTec AD equally alongside the five medical datasets. Default SMIA adapts all 12 Transformer blocks on $512\times 512$ medical and $768\times 768$ industrial canvases, applying paired training augmentations with probability $0.5$ (Supplementary Sec.~A.2). Same-backbone ablation and diagnostic configurations are detailed in \cref{sec:ablations} and Supplementary Sec.~C.1.

\subsection{Comparison with State-of-the-Art Methods}
\label{sec:main_results}

\Cref{tab:external} compares SMIA against six recent methods evaluated using their authors' official implementations under our fixed-support protocol (Supplementary Sec.~A.3): training-free VFM matchers INSID3~\cite{cuttano2026insid3} and RSRM~\cite{sun2026rsrm}, frozen-DINOv3 decoder learner DINO-AugSeg~\cite{xu2026dinoaugseg}, hierarchical SAM adapter DefectSAM~\cite{yan2025defectsam}, and CD-FSS methods HERA~\cite{ma2026hera} and MPA~\cite{nie2026mpa}. Per-category MVTec AD scores are reported in Supplementary Tab.~S1.

\paragraph{Overall and domain-level performance.}
SMIA achieves 77.40\% Overall mIoU at $N=5$ and 80.99\% at $N=10$, surpassing the strongest prior baseline at each budget—73.79\% for 5-shot HERA and 77.06\% for 10-shot DINO-AugSeg—by +3.61 and +3.93 percentage points (pp). Notably, ViT-S+/16 is smaller than the ViT-L/16, ViT-B/16, and ConvNeXt-L backbones used by INSID3, HERA, RSRM, and DINO-AugSeg, and SMIA updates only 0.51M parameters, or 1.7\% of DINO-AugSeg. These gains hold \emph{independently in both domains}. Across the five medical datasets, SMIA leads at both 5 shots (78.61\% vs.\ 77.46\%) and 10 shots (81.80\% vs.\ 79.60\%) with top foreground precision of 85.48\% and 85.93\% (Supplementary Tab.~S2), while widening its lead on MVTec AD to +4.98 and +5.38~pp. At $N=5$, SMIA also shows low draw-to-draw variance with a 0.28\% Overall SD, compared with 0.79\%--1.90\% for prior methods. It ranks first overall in every draw at both budgets, leads 16/20 and 18/20 tasks at 5 and 10 shots, and already exceeds the best 10-shot baseline mean using just 5 supports (77.40\% vs.\ 77.06\%).

Improvements are most pronounced where low-contrast lesions or thin defects resemble surrounding background texture. On medical benchmarks, SMIA gains +3.86 and +3.58~pp on ISIC 2018, improves 10-shot BUSI by +3.72~pp over DINO-AugSeg (70.32\% vs.\ 66.59\%), and reaches 94.38\% and 94.63\% on Chest X-ray. On MVTec AD, it leads 14 of the 15 categories at 10 shots, including margins of +12.31~pp on Screw and +7.25~pp on Grid. Prior baselines lead on only two 10-shot tasks: Kvasir-SEG, where HERA and RSRM reach 83.45\% and 83.00\% versus 82.21\% for SMIA, and MVTec Transistor, where DINO-AugSeg reaches 65.33\% versus 63.03\%.

\begin{table}[!tbp]
  \caption{\textbf{Industrial foreground precision--recall and mIoU breakdown across all 15 MVTec AD categories} (\%, five-draw mean; gains computed before rounding). \emph{Sup.-Free} denotes query-only inference without support inputs.}
  \label{tab:ind_pr}
  \centering
  {\setlength{\aboverulesep}{0pt}\setlength{\belowrulesep}{0pt}\renewcommand{\arraystretch}{1.15}\setlength{\tabcolsep}{3.3pt}%
  \resizebox{\linewidth}{!}{%
  \begin{tabular}{lcrrrrrr}
    \toprule
    \rowcolor{tblhead}
    & & \multicolumn{3}{c}{\textbf{5-Shot ($N=5$)}} & \multicolumn{3}{c}{\textbf{10-Shot ($N=10$)}} \\
    \cmidrule(lr){3-5} \cmidrule(lr){6-8}
    \rowcolor{tblhead}
    \textbf{Method} & \textbf{Sup.-Free} & \textbf{Prec.} & \textbf{Rec.} & \textbf{mIoU} & \textbf{Prec.} & \textbf{Rec.} & \textbf{mIoU} \\
    \midrule
    INSID3~\cite{cuttano2026insid3}      & \xmark & 26.39 & 47.82 & 56.75 & 26.78 & 52.76 & 57.98 \\
    RSRM~\cite{sun2026rsrm}              & \xmark & 17.22 & \secbest{69.85} & 52.62 & 16.67 & \textbf{72.74} & 52.40 \\
    HERA~\cite{ma2026hera}               & \xmark & 22.97 & \textbf{70.80} & 55.45 & 28.62 & \secbest{70.65} & 58.30 \\
    MPA~\cite{nie2026mpa}                & \xmark & 53.80 & 40.92 & 65.14 & 51.10 & 54.22 & 67.62 \\
    DefectSAM~\cite{yan2025defectsam}    & \cmark & \secbest{79.01} & 26.91 & 62.16 & \secbest{81.60} & 39.83 & 68.30 \\
    DINO-AugSeg~\cite{xu2026dinoaugseg}  & \cmark & 73.84 & 38.68 & \secbest{66.40} & 77.56 & 50.15 & \secbest{71.58} \\
    \rowcolor{oursbg}
    \textbf{SMIA (Ours)}                 & \cmark & \textbf{82.14} & 48.04 & \textbf{71.38}\,\gain{4.98} & \textbf{83.55} & 61.36 & \textbf{76.96}\,\gain{5.38} \\
    \bottomrule
  \end{tabular}%
  }}
  \vspace{-4pt}
\end{table}

\paragraph{Precision--recall behavior on industrial defects.}
Existing baselines split into two opposing regimes on MVTec AD (\cref{tab:ind_pr}). Support-retrieval matchers INSID3, RSRM, and HERA reach up to 72.74\% pixel recall, yet trigger widespread false positives across repetitive normal textures that drive foreground precision below 29\%. Conversely, support-free decoders DINO-AugSeg and DefectSAM maintain 73.84\%--81.60\% precision, but miss subtle defects with only 26.91\%--38.68\% recall at $N=5$ and 39.83\%--50.15\% at $N=10$. SMIA reaches 71.38\% and 76.96\% industrial mIoU by combining top precision (82.14\% and 83.55\%) with +9.35 and +11.20~pp higher recall than DINO-AugSeg. At inference time, it runs support-free on a single RTX 5090 in 13.7~ms on CVC-ClinicDB ($512^2$ canvas) and 23.5~ms on MVTec Bottle ($768^2$ canvas), using 6.7$\times$--8.9$\times$ less GPU memory than DINO-AugSeg (Supplementary Tab.~S3). Even on 10-shot Transistor, where DINO-AugSeg obtains higher mIoU (65.33\% vs.\ 63.03\%), SMIA maintains substantially higher foreground precision (78.22\% vs.\ 60.51\%).

\paragraph{Qualitative boundary discrimination.}
\Cref{fig:qualitative} compares predicted masks under shared 10-shot supports, with five additional cases in Supplementary Fig.~S1. On BUSI, RSRM misses the right lesion and DefectSAM over-segments surrounding stroma, whereas SMIA follows the lesion boundary closely. On Kvasir-SEG, MPA under-segments the polyp and DINO-AugSeg spills into adjacent mucosal folds, while SMIA delineates the full contour cleanly. On MVTec Bottle and Metal Nut, reference matchers activate on intact bottle cavities or brushed metal, and DINO-AugSeg and DefectSAM leave fragmented predictions; SMIA captures both the slender contaminant and the full oblique scratch.

\subsection{Same-Backbone and Structural Ablations}
\label{sec:ablations}

\begin{table}[!tbp]
  \caption{\textbf{Controlled same-backbone PEFT and subspace ablations on DINOv3 ViT-S+/16} ($N=10$, a shared fixed support--query split per task; gains vs.\ ParaX computed before rounding). \emph{w/o $\mathbf{B}_h$} omits off-diagonal mixing.}
  \label{tab:backbone}
  \centering
  {\setlength{\aboverulesep}{0pt}\setlength{\belowrulesep}{0pt}\renewcommand{\arraystretch}{1.13}\setlength{\tabcolsep}{4.0pt}%
  \resizebox{\linewidth}{!}{%
  \begin{tabular}{lrcccc}
    \toprule
    \rowcolor{tblhead}
    & & \multicolumn{3}{c}{\textbf{Dataset mIoU (\%)}} & \textbf{Summary} \\
    \cmidrule(lr){3-5} \cmidrule(lr){6-6}
    \rowcolor{tblhead}
    \textbf{Configuration} & \textbf{Params} & \textbf{MVTec} & \textbf{CVC} & \textbf{ISIC} & \textbf{Mean} \\
    \midrule
    Frozen + Linear                                       & 0.4K     & 74.98 & 74.90 & 77.95 & 75.94 \\
    AdaptFormer~\cite{chen2022adaptformer} ($b{=}55$)     & 512.5K   & 76.94 & 78.69 & 77.06 & 77.56 \\
    LoRA~\cite{hu2022lora} ($r{=}28,\alpha{=}56$)         & 516.5K   & 77.58 & 78.80 & 77.66 & 78.01 \\
    ParaX~\cite{lou2026parax} ($b{=}28$)                  & 507.7K   & 77.31 & 79.91 & 77.56 & 78.26 \\
    \midrule
    \rowcolor{ourslight}
    SMIA ($r{=}4$)                                        & 243.5K   & 78.12 & 80.09 & \textbf{79.52} & 79.24~\gain{0.99} \\
    \rowcolor{ourslight}
    SMIA ($r{=}8$, w/o $\mathbf{B}_h$)                    & 479.6K   & 78.01 & \secbest{80.90} & 79.24 & 79.38~\gain{1.12} \\
    \rowcolor{oursbg}
    \textbf{SMIA ($r{=}8$, default)}                      & \textbf{511.9K} & \secbest{78.23} & \secbest{80.90} & \secbest{79.33} & \secbest{79.48}~\gain{1.23} \\
    \rowcolor{ourslight}
    SMIA ($r{=}16$)                                       & 1,235.3K & \textbf{78.54} & \textbf{81.96} & 77.98 & \textbf{79.50}~\gain{1.24} \\
    \bottomrule
  \end{tabular}%
  }}
  \vspace{-4pt}
\end{table}

\begin{table}[!tbp]
  \caption{\textbf{Adaptation depth ablation across all 20 tasks ($N=10$, five draws, $r=8$).} \emph{Medical} and \emph{MVTec AD} average the 5 medical tasks and 15 industrial categories equally.}
  \label{tab:depth}
  \centering
  {\setlength{\aboverulesep}{0pt}\setlength{\belowrulesep}{0pt}\renewcommand{\arraystretch}{1.13}\setlength{\tabcolsep}{4.8pt}%
  \resizebox{\linewidth}{!}{%
  \begin{tabular}{lrccc}
    \toprule
    \rowcolor{tblhead}
    \textbf{Adapted Blocks} & \textbf{Params} & \textbf{Medical (5)} & \textbf{MVTec AD (15)} & \textbf{Overall mIoU} \\
    \midrule
    \rowcolor{ourslight}
    Last 4 Blocks  & 170.9K & 81.05\std{0.99} & 76.29\std{1.21} & 80.26\std{0.76} \\
    \rowcolor{ourslight}
    Last 8 Blocks  & 341.4K & \secbest{81.74}\std{1.22} & \secbest{76.70}\std{0.98} & \secbest{80.90}\std{1.05} \\
    \rowcolor{oursbg}
    \textbf{All 12 Blocks} & \textbf{511.9K} & \textbf{81.80}\bstd{1.24} & \textbf{76.96}\bstd{1.11} & \textbf{80.99}\bstd{1.10} \\
    \bottomrule
  \end{tabular}%
  }}
  \vspace{-4pt}
\end{table}

To separate adapter design from backbone and training differences, \cref{tab:backbone} compares frozen readout, three parameter-matched 12-block PEFT baselines, and SMIA on DINOv3 ViT-S+/16 across MVTec AD, CVC-ClinicDB, and ISIC 2018. The evaluated PEFT baselines comprise parallel FFN bottlenecks in AdaptFormer~\cite{chen2022adaptformer} at width $b=55$, query--value LoRA~\cite{hu2022lora} at rank $r=28$ and $\alpha=56$, and post-block multi-scale convolutions with image-routed shared experts in ParaX~\cite{lou2026parax} at width $b=28$. All adapted models initialize their linear head from the same frozen-feature fit on the 10 unaugmented supports and run 2,000 joint Adam steps under shared minibatches and augmentations (Supplementary Sec.~A.2).

The frozen linear probe yields 75.94\% mean mIoU, with 70.25\% foreground precision and 77.48\% recall. Among the evaluated PEFT baselines, AdaptFormer reaches 77.56\% mean mIoU, query--value LoRA reaches 78.01\%, and ParaX reaches 78.26\%. Although all three improve precision to 84.51\%--87.03\%, they lose 9.45--10.50~pp in recall (dropping to 66.99\%--68.03\%) and fall below the frozen probe on ISIC 2018 (77.06\%--77.66\% vs.\ 77.95\%). Keeping pretrained attention parameters frozen, default SMIA ($r=8$) raises precision from 70.25\% to 88.77\% while retaining higher recall (69.01\%) than all three PEFT baselines. Across the three benchmarks, it achieves mIoUs of 78.23\%, 80.90\%, and 79.33\%. Its 79.48\% mean mIoU surpasses frozen readout by +3.54~pp, AdaptFormer by +1.92~pp, LoRA by +1.47~pp, and ParaX by +1.23~pp. Even half-width SMIA ($r=4$, 243.5K parameters) exceeds every 0.51M PEFT baseline on all three datasets.

Within SMIA, diagonal pairing alone at $r=8$ already reaches 79.38\% mean mIoU (+1.12~pp over ParaX), while off-diagonal mixing $\mathbf{B}_h$ adds +0.22~pp on MVTec AD and +0.10~pp overall to reach 79.48\%. Scaling the subspace rank from $r=4$ (243.5K) to $r=16$ (1.24M) spans a narrow 79.24\%--79.50\% range, with $r=16$ edging $r=8$ by just +0.01~pp before rounding, indicating rapid rank saturation. Across all 20 tasks and five draws (\cref{tab:depth}), adapting the last 4, last 8, and all 12 blocks yields 80.26\%, 80.90\%, and 80.99\% Overall mIoU. Notably, adapting only the last 4 blocks already matches 12-block foreground recall (74.79\% vs.\ 74.80\%), whereas extending SMIA into earlier blocks steadily lifts precision from 83.62\% to 85.12\% and 85.54\% (Supplementary Sec.~B.2).

\subsection{Mechanistic Analysis: State-Driven Updates}
\label{sec:mechanism}

\begin{figure}[!tbp]
  \centering
  \includegraphics[width=0.86\linewidth]{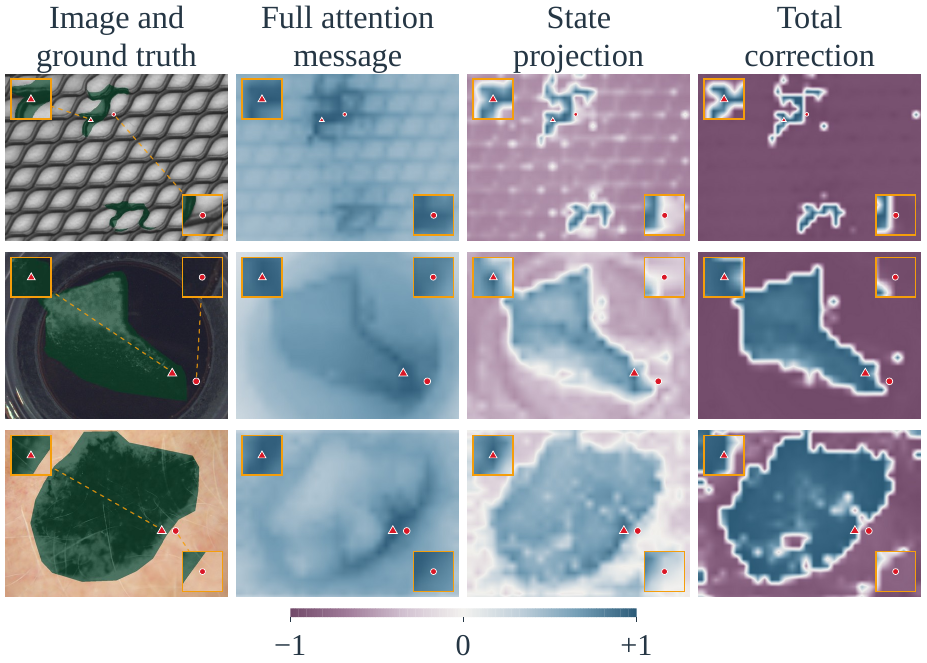}
  \caption{\textbf{Opposing corrections from strongly aligned boundary messages.} Rows show MVTec Grid, MVTec Bottle, and ISIC 2018 queries ($N=10$, Block~12). Green overlays mark ground truth; red triangles ($A$) and circles ($B$) mark nearby foreground and background boundary patches (amber zoom insets). Right three columns plot cosine similarity to $A$ ($[-1, +1]$) for concatenated six-head message, state projection, and total correction; aligned messages at $A$ and $B$ map to opposing updates.}
  \label{fig:direction}
  \vspace{-5pt}
\end{figure}

\begin{figure*}[!tbp]
  \centering
  \includegraphics[width=0.73\textwidth]{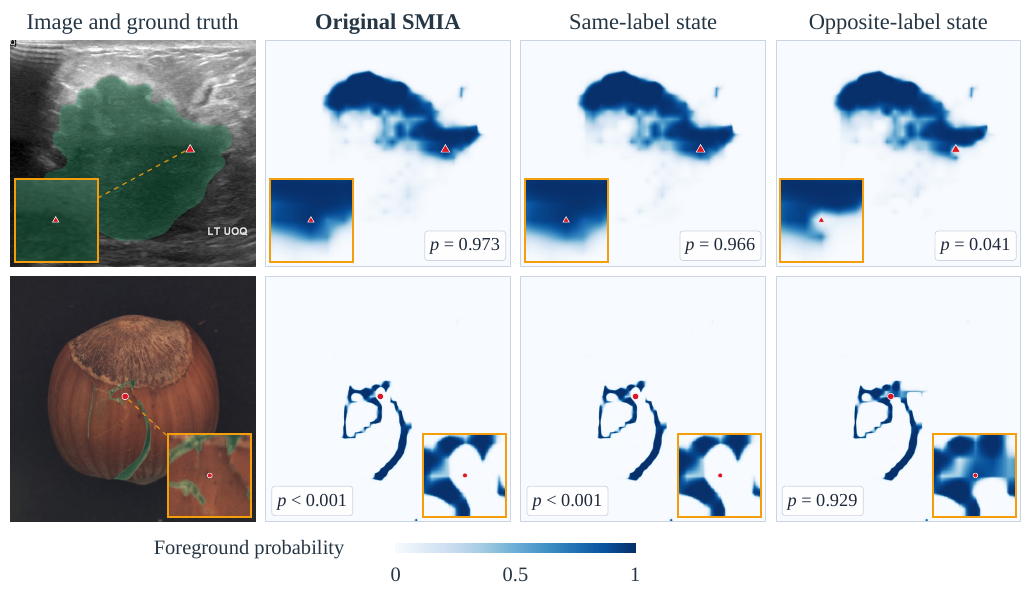}
  \caption{\textbf{Counterfactual state-swap intervention under fixed attention messages.} Rows show BUSI lesion ($A$, red triangle) and MVTec Hazelnut intact shell ($B$, red circle), with amber zoom insets and restored pixel probabilities $p$; heatmaps aggregate independent per-patch swaps across matched positions. Holding Block-12 features, six-head messages, and correction $\ell_2$-norm fixed, swapping \emph{only} SMIA's state projection from opposite-label donors ($d \le 0.20$) flips local predictions.}
  \label{fig:swap}
  \vspace{-4pt}
\end{figure*}

\paragraph{Divergent updates under aligned boundary messages.}
\Cref{fig:direction} examines Block~12 of 12-block SMIA at nearby foreground ($A$) and background ($B$) boundary patches on MVTec Grid, Bottle, and ISIC 2018. Across these boundaries, $A$ and $B$ receive positively aligned concatenated six-head messages with cosines of 0.943, 0.868, and 0.829, whereas their local state projections separate clearly at $-0.226$, $-0.083$, and $0.242$. Weighting the shared basis matrices by local states in \cref{eq:operator_bank} sends the resulting total corrections $\boldsymbol{\delta}_A$ and $\boldsymbol{\delta}_B$ in opposite directions, reaching cosines of $-0.998$, $-0.977$, and $-0.836$. Eight additional boundary pairs in Supplementary Fig.~S4 exhibit the same behavior, mapping message cosines in $[0.766, 0.944]$ to correction cosines in $[-0.997, -0.645]$. By contrast, unadapted terminal features group tokens by part identity: on MVTec Cable, damaged inner insulation shares higher frozen cosine with intact inner insulation than with outer-sheath cuts (0.707 vs.\ 0.552; Supplementary Sec.~C.1).

\paragraph{Causal verification via norm-matched state swapping.}
Because both messages and state projections vary between $A$ and $B$ in \cref{fig:direction}, we conduct a fixed-message intervention at Block~12 (\cref{fig:swap}; Supplementary Sec.~C.2 and Fig.~S3) to isolate the effect of the state projections. For the two cases in \cref{fig:swap}, we freeze the target patch's post-attention feature and six-head message, and swap \emph{only} SMIA's state projection from a pure-foreground or pure-background donor patch on the \emph{same} donor query image, matched within message cosine distance $\le 0.20$ and norm ratio $[0.8, 1.25]$. Let $\boldsymbol{\delta}'$ denote the recomputed update before rescaling. We rescale it to $\|\boldsymbol{\delta}\|_2$ and add $\boldsymbol{\delta}' \cdot (\|\boldsymbol{\delta}\|_2 / \|\boldsymbol{\delta}'\|_2)$ to the frozen post-attention feature, changing only the update direction while keeping its magnitude fixed.

In \cref{fig:swap}, equal-norm same-label swaps preserve local predictions on both BUSI and MVTec Hazelnut. Equal-norm opposite-label swaps rotate the BUSI lesion update to $\cos = 0.033$, flipping its patch logit from $+3.856$ to $-3.345$ and restored pixel probability from 0.97 to 0.04. On Hazelnut, they invert the intact-shell update at $\cos = -0.999$, flipping its patch logit from $-24.921$ to $+2.789$ and pixel probability from $<$0.001 to 0.93. Scaling magnitude alone via $\|\boldsymbol{\delta}'\|_2 / \|\boldsymbol{\delta}\|_2 \cdot \boldsymbol{\delta}$ flips neither prediction, confirming that SMIA uses local state directionally to steer local decisions.

%% file: sec/5_conclusion.tex
\section{Conclusion and Discussion}
\label{sec:conclusion}

We presented the State--Message Interaction Adapter (SMIA) for cross-domain few-shot segmentation on vision foundation models. Keeping pretrained attention and FFN parameters frozen, SMIA conditions each head's post-aggregation message update on the receiving patch's pre-attention state within low-dimensional subspaces, before frozen FFNs and deeper blocks refine the updated features for linear readout. With only 0.51M trainable parameters, SMIA achieves six-dataset mean mIoUs of 77.40\% at 5 shots and 80.99\% at 10 shots across 20 medical and industrial tasks, exceeding the strongest prior method at each budget by +3.61 and +3.93 percentage points. On one RTX 5090, single-pass, support-free inference takes 13.7~ms (166~MiB) on CVC-ClinicDB and 23.5~ms (220~MiB peak PyTorch-allocated GPU memory) on MVTec Bottle. Controlled 10-shot comparisons across three benchmarks on the same DINOv3 backbone further show higher mIoU and a better precision--recall balance than parameter-matched AdaptFormer, LoRA, and ParaX. Forward boundary comparisons reveal opposing updates for closely aligned messages, and norm-matched state swaps under fixed messages flip local predictions. Together, these results support learning how per-head messages update each receiving representation as a function of local state while keeping pretrained attention parameters frozen.

\paragraph{Limitations and future work.}
SMIA conditions each head's update on the receiving patch state without explicit cross-region geometric constraints. While effective on localized lesions and surface defects, on MVTec Transistor its recall remains conservative for missing or displaced leads—where exposed board regions can appear locally intact—despite 78.22\% foreground precision. Combining local state-conditioned updates with non-local relational priors to capture global part geometry is a promising direction for future work.

%% file: sec/X_suppl.tex
\appendix
\renewcommand{\thetable}{S\arabic{table}}
\renewcommand{\thefigure}{S\arabic{figure}}
\renewcommand{\theequation}{S\arabic{equation}}
\setcounter{table}{0}
\setcounter{figure}{0}
\setcounter{equation}{0}
\setcounter{dbltopnumber}{2}

\noindent\textit{\Cref{sec:supp_repro} details data preparation, SMIA calibration and optimization, and external baseline configurations. \Cref{sec:supp_results} compares segmentation results across tasks and support budgets, reports depth and width ablations, compares inference costs, and evaluates the one-shot setting. \Cref{sec:supp_mechanism} examines frozen patch features, state-swap interventions, and update directions across target boundaries.}

\section{Reproducibility Details}
\label{sec:supp_repro}

\begin{table*}[!tbp]
  \caption{\textbf{Per-category MVTec AD comparison under $N=5$ and $N=10$ support budgets.} Values report native-resolution mIoU (\%, mean $\pm$ SD across five nested random support draws; gains computed before rounding). Best and second-best results per category and budget are \textbf{bolded} and \secbest{underlined}.}
  \label{tab:mvtec_per_class}
  \vspace{-4pt}
  \centering
  {\setlength{\aboverulesep}{0pt}\setlength{\belowrulesep}{0pt}\renewcommand{\arraystretch}{1.04}\setlength{\tabcolsep}{4.6pt}%
  \resizebox{\textwidth}{!}{%
  \begin{tabular}{lcccccc>{\columncolor{oursbg}}c}
    \toprule
    \rowcolor{tblhead}
    \textbf{MVTec Category} & \textbf{DefectSAM}~\cite{yan2025defectsam} & \textbf{DINO-AugSeg}~\cite{xu2026dinoaugseg} & \textbf{HERA}~\cite{ma2026hera} & \textbf{INSID3}~\cite{cuttano2026insid3} & \textbf{RSRM}~\cite{sun2026rsrm} & \textbf{MPA}~\cite{nie2026mpa} & \textbf{SMIA (Ours)} \\
    \midrule
    \rowcolor{tblsub}
    \multicolumn{8}{l}{\textit{\textbf{5-Shot Support Budget ($N = 5$)}}} \\
    \midrule
    Bottle      & 72.12\std{7.15}  & 75.49\std{4.61}  & 63.74\std{14.54} & 55.51\std{11.63} & 49.61\std{5.28} & \secbest{78.39}\std{3.08}  & \textbf{81.96}\bstd{6.47} \\
    Cable       & 64.44\std{5.18}  & \secbest{66.22}\std{3.49}  & 51.61\std{8.68}  & 54.58\std{9.31}  & 49.66\std{8.71} & 63.29\std{5.74}  & \textbf{68.72}\bstd{8.12} \\
    Capsule     & 52.94\std{2.96}  & \secbest{57.09}\std{5.11}  & 52.62\std{0.88}  & 56.24\std{1.16}  & 43.16\std{0.76} & 53.46\std{0.85}  & \textbf{59.30}\bstd{3.91} \\
    Carpet      & 56.57\std{6.90}  & 73.88\std{6.10}  & 71.13\std{6.84}  & 56.30\std{5.71}  & \secbest{75.92}\std{6.13} & 71.10\std{3.74}  & \textbf{78.88}\bstd{5.14} \\
    Grid        & 53.98\std{1.70}  & 64.72\std{1.49}  & 63.18\std{2.56}  & \secbest{70.19}\std{0.84}  & 62.38\std{1.46} & 66.42\std{2.74}  & \textbf{73.25}\bstd{2.99} \\
    Hazelnut    & \secbest{66.32}\std{6.88}  & 62.61\std{4.03}  & 59.08\std{4.44}  & 54.67\std{3.02}  & 40.97\std{1.61} & 59.28\std{4.38}  & \textbf{69.48}\bstd{7.99} \\
    Leather     & 64.91\std{6.03}  & 63.82\std{6.67}  & 58.58\std{6.13}  & 54.06\std{0.56}  & 55.75\std{1.22} & \secbest{68.85}\std{5.12}  & \textbf{75.21}\bstd{5.27} \\
    Metal Nut   & \secbest{81.25}\std{17.96} & 78.80\std{16.86} & 37.78\std{10.15} & 44.52\std{8.72}  & 39.78\std{8.15} & 77.47\std{15.29} & \textbf{84.58}\bstd{17.78} \\
    Pill        & \secbest{55.68}\std{12.62} & 53.68\std{3.49}  & 46.73\std{12.06} & 51.66\std{2.05}  & 43.37\std{6.73} & 52.76\std{3.14}  & \textbf{59.23}\bstd{13.43} \\
    Screw       & 51.46\std{0.65}  & \secbest{56.10}\std{2.06}  & 46.98\std{6.16}  & 53.14\std{2.24}  & 49.03\std{0.35} & 53.98\std{2.45}  & \textbf{64.65}\bstd{6.46} \\
    Tile        & 76.85\std{3.30}  & 77.15\std{5.61}  & 65.97\std{9.96}  & 75.78\std{2.86}  & 80.52\std{4.38} & \textbf{83.86}\bstd{5.23} & \secbest{81.28}\std{4.70} \\
    Toothbrush  & 56.56\std{6.19}  & \secbest{66.81}\std{5.50}  & 52.98\std{2.64}  & 56.92\std{5.65}  & 41.88\std{1.02} & 60.79\std{6.54}  & \textbf{66.88}\bstd{7.65} \\
    Transistor  & 51.59\std{1.17}  & \textbf{60.28}\bstd{5.49} & 37.08\std{10.53} & 48.20\std{2.36}  & 46.72\std{3.47} & 52.19\std{1.19}  & \secbest{58.05}\std{4.81} \\
    Wood        & \secbest{68.41}\std{6.57}  & 66.89\std{8.15}  & 61.80\std{11.92} & 64.86\std{7.44}  & 64.81\std{4.00} & 63.45\std{5.11}  & \textbf{74.56}\bstd{4.99} \\
    Zipper      & 59.28\std{3.58}  & \secbest{72.46}\std{4.21}  & 62.44\std{10.33} & 54.62\std{6.92}  & 45.73\std{0.56} & 71.85\std{4.91}  & \textbf{74.76}\bstd{4.60} \\
    \midrule
    \rowcolor{ourslight}
    \textbf{15-Class Mean} & 62.16\std{0.73} & \secbest{66.40}\std{1.72} & 55.45\std{2.26} & 56.75\std{1.76} & 52.62\std{0.95} & 65.14\std{0.48} & \textbf{71.38}\bstd{1.70}~\gain{4.98} \\
    \midrule
    \rowcolor{tblsub}
    \multicolumn{8}{l}{\textit{\textbf{10-Shot Support Budget ($N = 10$)}}} \\
    \midrule
    Bottle      & 80.49\std{4.99}  & 78.40\std{3.91}  & 63.14\std{14.27} & 66.48\std{1.50}  & 51.10\std{3.74} & \secbest{81.25}\std{2.16}  & \textbf{86.20}\bstd{4.31} \\
    Cable       & 69.06\std{3.48}  & \secbest{71.39}\std{3.19}  & 70.22\std{4.65}  & 55.30\std{7.05}  & 47.38\std{4.69} & 65.47\std{2.73}  & \textbf{77.75}\bstd{1.43} \\
    Capsule     & 54.77\std{3.04}  & \secbest{58.62}\std{4.27}  & 55.03\std{1.12}  & 55.70\std{1.33}  & 42.31\std{1.31} & 54.91\std{1.57}  & \textbf{61.94}\bstd{4.53} \\
    Carpet      & 65.15\std{3.85}  & \secbest{78.24}\std{1.22}  & 74.50\std{0.78}  & 53.74\std{1.71}  & 75.63\std{2.42} & 73.49\std{1.63}  & \textbf{81.92}\bstd{1.83} \\
    Grid        & 62.73\std{3.00}  & 68.17\std{1.67}  & 65.84\std{1.25}  & \secbest{69.85}\std{1.65}  & 62.96\std{1.34} & 67.07\std{0.76}  & \textbf{77.11}\bstd{0.78} \\
    Hazelnut    & 72.60\std{5.88}  & \secbest{74.62}\std{7.23}  & 54.35\std{13.12} & 53.83\std{1.32}  & 41.35\std{0.55} & 65.53\std{1.26}  & \textbf{79.63}\bstd{2.58} \\
    Leather     & \secbest{72.52}\std{2.68}  & 72.33\std{2.79}  & 55.08\std{0.98}  & 54.54\std{0.34}  & 55.05\std{1.77} & 71.51\std{2.49}  & \textbf{78.36}\bstd{3.12} \\
    Metal Nut   & 83.47\std{18.02} & \secbest{84.16}\std{17.65} & 58.07\std{16.98} & 45.93\std{3.52}  & 38.48\std{6.51} & 75.57\std{12.73} & \textbf{86.48}\bstd{18.03} \\
    Pill        & 66.81\std{16.28} & \secbest{67.47}\std{14.92} & 42.20\std{11.66} & 51.45\std{2.51}  & 41.08\std{8.40} & 61.27\std{10.27} & \textbf{69.97}\bstd{15.31} \\
    Screw       & 54.00\std{2.79}  & \secbest{60.20}\std{2.99}  & 49.19\std{5.62}  & 53.73\std{0.83}  & 48.50\std{0.21} & 52.97\std{1.03}  & \textbf{72.50}\bstd{2.49} \\
    Tile        & 84.16\std{3.39}  & 84.22\std{2.47}  & 59.92\std{3.15}  & 79.41\std{2.09}  & 81.00\std{5.04} & \secbest{87.53}\std{0.92}  & \textbf{87.72}\bstd{3.05} \\
    Toothbrush  & 61.11\std{6.08}  & 65.15\std{7.22}  & 61.58\std{1.58}  & 58.64\std{2.93}  & 41.98\std{1.04} & \secbest{66.03}\std{2.27}  & \textbf{72.46}\bstd{4.50} \\
    Transistor  & 55.29\std{3.10}  & \textbf{65.33}\bstd{4.91} & 40.14\std{4.41} & 46.43\std{4.82}  & 47.02\std{2.81} & 51.72\std{2.40}  & \secbest{63.03}\std{5.11} \\
    Wood        & \secbest{73.20}\std{4.23}  & 69.78\std{4.77}  & 66.22\std{2.95}  & 69.06\std{6.19}  & 66.77\std{1.76} & 66.93\std{2.95}  & \textbf{77.49}\bstd{3.14} \\
    Zipper      & 69.18\std{3.42}  & \secbest{75.62}\std{2.01}  & 58.97\std{15.82} & 55.65\std{4.67}  & 45.35\std{0.24} & 73.09\std{3.02}  & \textbf{81.83}\bstd{0.84} \\
    \midrule
    \rowcolor{ourslight}
    \textbf{15-Class Mean} & 68.30\std{0.51} & \secbest{71.58}\std{1.51} & 58.30\std{1.52} & 57.98\std{1.41} & 52.40\std{0.80} & 67.62\std{0.48} & \textbf{76.96}\bstd{1.11}~\gain{5.38} \\
    \bottomrule
  \end{tabular}%
  }}
\end{table*}

\subsection{Dataset Processing, Splits, and Scoring}
\label{sec:supp_data}

Following the 20-task evaluation protocol under $N \in \{5, 10\}$ support budgets in Main Paper Sec.~4.1, all compared methods share the same image inventories and are assigned identical nested support sets. BUSI and MVTec AD enforce $3/2$ and $6/4$ positive-to-negative support quotas at $N=5$ and $N=10$, backfilling from the other class only when fewer eligible scans exist. Every scan in Kvasir-SEG, CVC-ClinicDB, ISIC 2018 Task~1, and Chest X-ray contains foreground targets; for Chest X-ray, we include only the 704 radiographs with paired lung-field masks, comprising 566 from Shenzhen and 138 from Montgomery. In the $N=1$ evaluation (\cref{sec:supp_oneshot}), we examine each task's nested 5-shot support set in its original order and select the first foreground image whose downsampled mask meets the feature-grid reference criteria in \cref{sec:supp_baselines}. All seven methods share this image as their sole support and evaluate on all remaining inventory images as queries.

To construct binary ground-truth masks, we take the pixelwise union when a BUSI ultrasound scan has multiple lesion mask files; for Kvasir-SEG, CVC-ClinicDB, ISIC 2018 Task~1, and Chest X-ray, we label grayscale mask pixels with intensity $\ge 128$ as foreground to remove boundary anti-aliasing and compression artifacts; for MVTec AD, we mark nonzero defect-mask pixels as foreground and assign all-background masks to normal images. For each task, we accumulate pixel confusion counts across all query images at native resolution before computing foreground IoU, background IoU, mIoU, foreground precision, and foreground recall; if a metric denominator is zero, IoU and recall default to $1$ and precision defaults to $0$.

\subsection{Calibration, Optimization, and Augmentations}
\label{sec:supp_train}

\begin{figure*}[!tbp]
  \centering
  \includegraphics[width=0.96\textwidth]{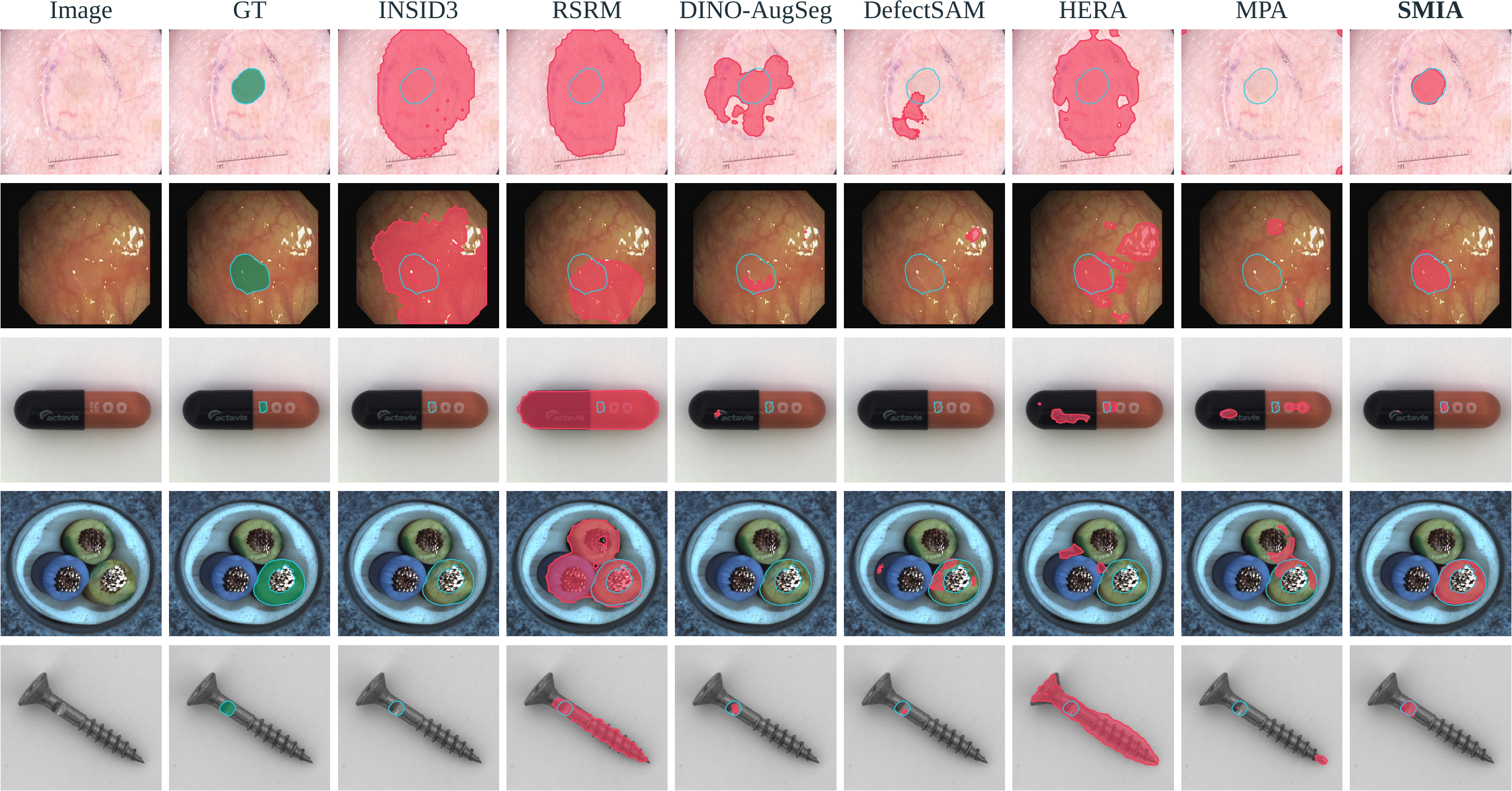}
  \vspace{-4pt}
  \caption{\textbf{Additional qualitative comparisons across medical and industrial tasks ($N=10$).} The first two rows show an ISIC 2018 skin lesion and a CVC-ClinicDB polyp; the last three rows show a faulty imprint on a capsule, swapped cable cores, and a scratch on the neck of a screw. The GT column overlays ground truth in green, while method columns show cyan ground-truth contours and coral-red predicted foreground under the same 10-shot support split per row.}
  \label{fig:supp_qual}
  \vspace{-7pt}
\end{figure*}

\paragraph{Area-weighted RMS product calibration.}
At initialization (Main Paper Sec.~3.2), rows of $\mathbf{S}_h \in \R^{r \times d}$ and $\mathbf{M}_h \in \R^{r \times d_h}$ are sampled from $\mathcal{N}(\mathbf{0}, \mathbf{I})$ and normalized to unit $\ell_2$ norm, with $\mathbf{B}_h = \mathbf{0}$ and $\mathbf{U}_h = \mathbf{0}$. Wider ablations with $r > 8$ retain the first eight rows of the $r=8$ initialization and append independently sampled unit-norm Gaussian rows. Because aspect-ratio-preserving resize introduces variable zero-padded borders across scans, SMIA estimates $\mathbf{C}_h \in \R^{r \times r}$ from the $N$ unaugmented supports before optimization while weighting each patch by its unpadded area fraction, so padded borders and larger scans do not bias the scale. Let $\omega_{nt} \ge 0$ denote the unpadded area fraction of patch~$t$ on support~$n$, normalized so $\sum_{t=1}^{N_p} \omega_{nt} = 1$. Entry $c_{hij}$ is given by
\begin{equation}
  c_{hij} = \sqrt{\frac{1}{N}\sum_{n=1}^{N} \sum_{t=1}^{N_p} \omega_{nt} \left(a_{nthi}\, b_{nthj}\right)^2},
  \label{eq:supp_rms}
\end{equation}
We replace any zero entry with $1$ and keep $\mathbf{C}_h$ fixed throughout training.

\paragraph{Solver schedule and same-backbone baselines.}
Before joint optimization (Main Paper Sec.~3.4), we pre-fit the 385-parameter linear head $(\mathbf{w}_{\mathrm{cls}}, b_{\mathrm{cls}})$ on frozen unaugmented support features by minimizing native-resolution class-balanced BCE with ridge penalty $\frac{\lambda}{2}(\|\mathbf{w}_{\mathrm{cls}}\|_2^2 + b_{\mathrm{cls}}^2)$ at $\lambda = 10^{-3}$. We solve this convex objective via L-BFGS followed by matrix-free Newton--CG until the upper bound $\|\nabla\mathcal{L}_{\mathrm{init}}\|_2^2/(2\lambda)$ on the objective gap is at most $10^{-8}$. We then drop the ridge term and jointly optimize all adapter and readout parameters for 2,000 Adam steps at learning rate $10^{-4}$ with $\beta_1=0.9$, $\beta_2=0.999$, $\epsilon=10^{-8}$, and zero weight decay. At $N=5$ and $N=10$, each step samples two distinct supports without replacement and applies the unit-sphere retraction in Main Paper Eq.~(6) to all rows of $\mathbf{S}_h$ and $\mathbf{M}_h$. At $N=1$, SMIA keeps the same 12-block width-8 architecture and 2,000-step schedule, calibrating $\mathbf{C}_h$ and pre-fitting the linear head on the single unaugmented support before sampling two independent paired-augmented views of that image--mask pair per step.

In Main Paper Tab.~3, every adapted baseline starts from this identical pre-fit head, shares the same minibatch sequence and paired augmentations over 2,000 Adam steps, and is evaluated directly at step 2,000 under task-shared hyperparameters. AdaptFormer~\cite{chen2022adaptformer} inserts a width-55 ReLU bottleneck in parallel with the FFN in each of the 12 blocks (scale $0.1$, dropout $0.1$, zero-initialized up-projection), totaling $512{,}533$ parameters. LoRA~\cite{hu2022lora} applies rank-28 updates ($\alpha=56$) to both the query and value projections in each of the 12 blocks for $516{,}481$ parameters. ParaX~\cite{lou2026parax} places a width-28 patch adapter with $3{\times}3$, $5{\times}5$, and $7{\times}7$ depthwise kernels after each of the 12 blocks, sharing a bank of 12 experts with a per-block router of hidden width $16$ ($507{,}733$ trainable parameters).

\paragraph{Canvas geometry and paired support augmentations.}
Medical and MVTec AD images are resized with preserved aspect ratio and center-padded onto $512\times 512$ and $768\times 768$ canvases, with padded borders zero-filled after ImageNet RGB normalization. We bilinearly upsample the $32\times 32$ or $48\times 48$ patch-grid logits to the padded canvas, remove the padding, and resize them to the native image dimensions (both steps using \texttt{align\_corners=False}); border pixels rotated into the frame carry zero loss weight. Each time a support image is sampled during training, we apply paired augmentation with probability $0.5$: when active, brightness and contrast are scaled independently by factors in $[0.9, 1.1]$, while $3\times 3$ Gaussian blur ($\sigma \in [0.1, 0.5]$), hue jitter ($[-0.01, 0.01]$), and horizontal flips on the five medical tasks are each applied independently with probability $0.5$. Rotation uses $90^\circ/180^\circ/270^\circ$ quarter-turns with probability $0.25$ on orientation-unconstrained top-view parts (Metal Nut, Screw, and Hazelnut), and continuous angles in $[-5^\circ, 5^\circ]$ otherwise. Calibration, head pre-fitting, and test evaluation use unaugmented images.

\subsection{External Baseline Configurations}
\label{sec:supp_baselines}

In Main Paper Tab.~1 and \cref{tab:mvtec_per_class}, DINO-AugSeg~\cite{xu2026dinoaugseg}, DefectSAM~\cite{yan2025defectsam}, and MPA~\cite{nie2026mpa} train on all $N$ support image--mask pairs per draw, including normal all-background masks. HERA~\cite{ma2026hera} fits once per draw over $M$ eligible positive supports for $M-1$ Adam steps with held-out and subset cross-entropy regularization, enumerating all subsets when at most six exist and sampling six distinct subsets otherwise, before reusing the adapted model across all queries. Training-free matchers INSID3~\cite{cuttano2026insid3} and RSRM~\cite{sun2026rsrm} match query features directly against positive supports at test time (HERA and RSRM require nonzero foreground and background regions on the downsampled feature grid, with at least two tokens per class for RSRM).

All six baselines build on their public implementations, taking architectures, objectives, and optimizers from their papers and released configurations under our fixed-support protocol. INSID3, RSRM, and HERA use $1024^2$, $480^2$, and $400^2$ inputs, respectively. DefectSAM fine-tunes SAM-B for 150 AdamW epochs at learning rate $2\times 10^{-4}$ on long-edge-padded $1024^2$ inputs. MPA uses its source-free ImageNet ResNet-50 implementation with refinement at learning rate $5\times 10^{-4}$ and $400^2$ inputs, running 40 SGD updates per support set ($3$ in stages 1--5 and $25$ in stage 6); we evaluate the final checkpoint and supply the support images and masks at query time. DINO-AugSeg trains its $30{,}332{,}418$ parameters with Adam at learning rate $10^{-4}$ for 250 epochs at $224^2$ on ISIC 2018 and 1,000 epochs at $512^2$ elsewhere, evaluating final-epoch weights. In \cref{tab:supp_oneshot} ($N=1$), HERA and MPA use their official one-shot branches: HERA synthesizes two soft copy-paste views alongside the support for two Adam steps at learning rate $1.3\times 10^{-3}$, MPA constructs six views from the support for its 40-step SGD schedule, and DINO-AugSeg and DefectSAM train with batch size $1$. At inference, RSRM, DINO-AugSeg, HERA, and MPA apply two-class argmax and nearest-neighbor resize to native size, whereas INSID3 and DefectSAM output native-size binary masks directly.

\section{Extended Benchmark and Ablation Analysis}
\label{sec:supp_results}

\subsection{Per-Category MVTec AD and Domain Analysis}
\label{sec:supp_mvtec}

Beyond Main Paper Tab.~1, we group the 15 MVTec AD categories in \cref{tab:mvtec_per_class} into five texture categories (carpet, grid, leather, tile, and wood) and ten object categories. Anomalous pixels cover only $0.08\%$ to $3.87\%$ of image area across MVTec AD and fall below $1.0\%$ in nine categories. On textures, where intact patterns recur across scans, matchers INSID3, RSRM, and HERA reach $36.60\%$--$41.00\%$ 10-shot foreground precision and $64.31\%$--$68.28\%$ mIoU. On object categories, their 10-shot precision drops to $4.80\%$--$24.63\%$ and mIoU to $44.45\%$--$55.29\%$, as matching can also highlight intact structures (\cref{fig:supp_qual}). By contrast, support-free decoders DINO-AugSeg and DefectSAM keep $77.76\%$--$78.66\%$ 10-shot foreground precision on object categories, but their mean recall stays at $24.64\%$--$34.96\%$ ($N=5$) and $36.46\%$--$45.79\%$ ($N=10$).

\begin{figure*}[!tbp]
  \centering
  \includegraphics[width=0.96\textwidth]{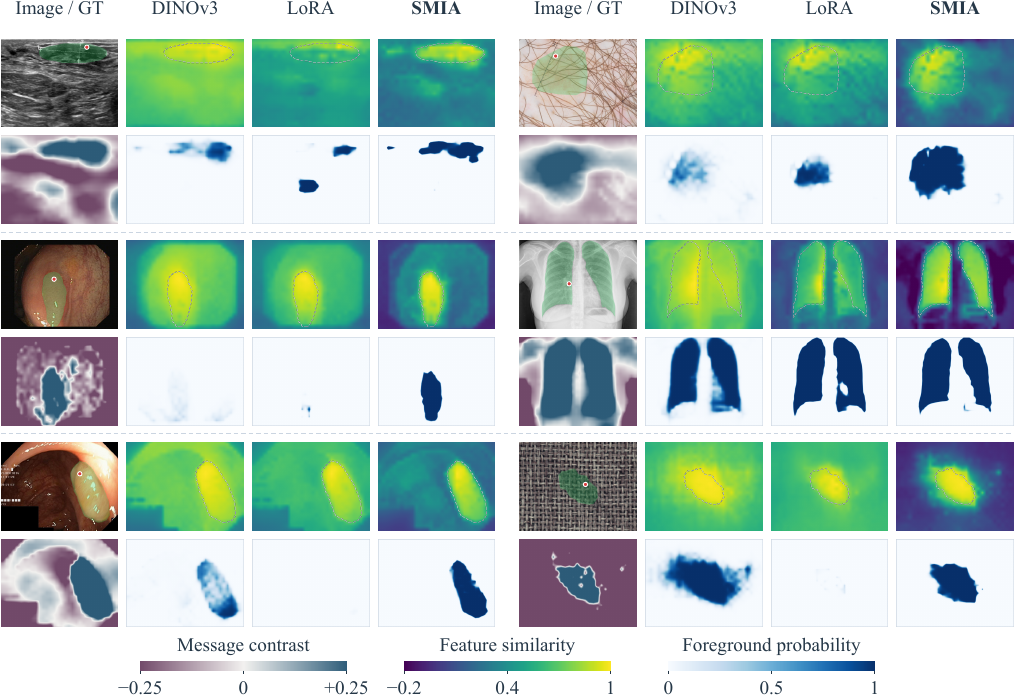}
  \vspace{-4pt}
  \caption{\textbf{Extended representation and segmentation diagnostics across six tasks ($N=10$).} Left, top to bottom: BUSI, CVC-ClinicDB, and Kvasir-SEG; right: ISIC 2018, Chest X-ray, and MVTec AD Carpet. In each two-row case, column~1 shows the query image with green ground truth and a red reference patch above and frozen single-head message contrast below (\cref{eq:supp_contrast}; teal/plum indicate higher mean similarity to foreground/nearby background references). Columns 2--4 show frozen DINOv3, 12-block Q/V LoRA ($r=8$), and 4-block SMIA ($r=8$), plotting final-layer cosine similarity to the red patch with dashed white ground-truth contours above and foreground probabilities from linear heads fit on the same 10 supports below (\cref{sec:supp_diag}).}
  \label{fig:supp_motivation}
  \vspace{-7pt}
\end{figure*}

\begin{figure*}[!tbp]
  \centering
  \includegraphics[width=0.96\textwidth]{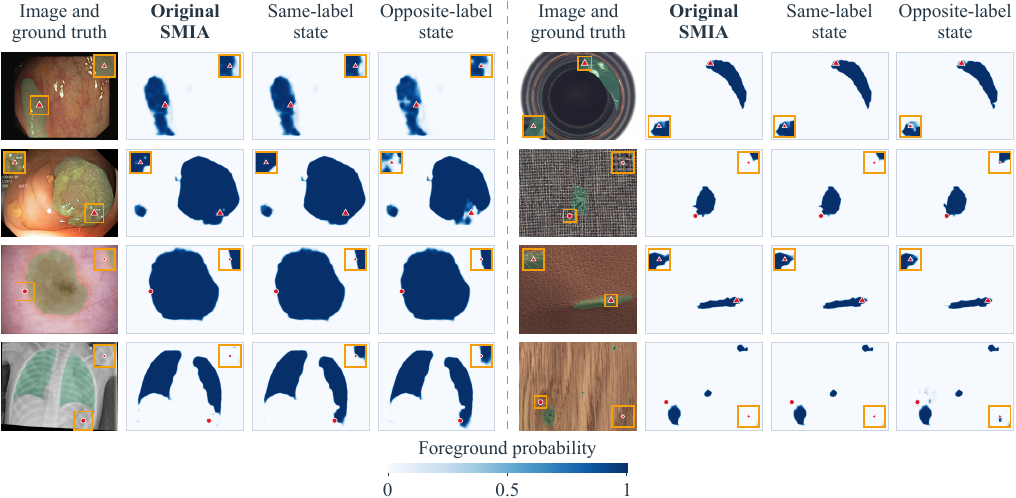}
  \vspace{-4pt}
  \caption{\textbf{Extended counterfactual state-swap interventions across medical and industrial tasks ($N=10$).} Left, top to bottom: CVC-ClinicDB, Kvasir-SEG, ISIC 2018, and Chest X-ray; right: MVTec Bottle, Carpet, Leather, and Wood. Column~1 shows the query with green ground truth, and columns 2--4 show foreground probabilities in $[0, 1]$ for original SMIA, same-label swaps, and opposite-label swaps. Red triangles mark four foreground patches, red circles mark four background patches, and amber insets magnify them. The swap maps combine independent interventions at matched patches ($d \le 0.30$ on CVC-ClinicDB, Kvasir-SEG, and Chest X-ray; $d \le 0.20$ elsewhere), with both swaps preserving the original correction norm; at all eight marked patches, same-label swaps preserve the predicted class and opposite-label swaps flip it (\cref{sec:supp_swap}).}
  \label{fig:supp_swap}
  \vspace{-7pt}
\end{figure*}

SMIA leads mean mIoU in both industrial groups, ranking first on 13 of 15 MVTec AD categories at $N=5$ and 14 of 15 at $N=10$ (\cref{tab:mvtec_per_class}). On textures, it reaches 76.64\% at 5-shot and 80.52\% at 10-shot, surpassing 5-shot MPA ($70.74\%$) by $+5.90$~pp and 10-shot DINO-AugSeg ($74.55\%$) by $+5.97$~pp. On object categories, SMIA reaches 68.76\% and 75.18\% mIoU, ahead of DINO-AugSeg by $+3.80$ and $+5.08$~pp, reaching $56.02\%$ 10-shot foreground recall (vs.\ $45.79\%$ for DINO-AugSeg) alongside $84.48\%$ precision. At $N=10$, Transistor is the sole MVTec AD category where DINO-AugSeg exceeds SMIA in mIoU ($65.33\%$ vs.\ $63.03\%$), while SMIA keeps higher foreground precision ($78.22\%$ vs.\ $60.51\%$).

\begin{table}[!tbp]
  \caption{\textbf{Foreground precision, recall, and mIoU averaged equally across the five medical datasets} (\%, five-draw mean; gains computed before rounding). \emph{Sup.-Free} denotes query-only inference without support inputs.}
  \label{tab:med_pr}
  \centering
  {\setlength{\aboverulesep}{0pt}\setlength{\belowrulesep}{0pt}\renewcommand{\arraystretch}{1.15}\setlength{\tabcolsep}{3.8pt}%
  \resizebox{\linewidth}{!}{%
  \begin{tabular}{lcrrrrrr}
    \toprule
    \rowcolor{tblhead}
    & & \multicolumn{3}{c}{\textbf{5-Shot ($N=5$)}} & \multicolumn{3}{c}{\textbf{10-Shot ($N=10$)}} \\
    \cmidrule(lr){3-5} \cmidrule(lr){6-8}
    \rowcolor{tblhead}
    \textbf{Method} & \textbf{Sup.-Free} & \textbf{Prec.} & \textbf{Rec.} & \textbf{mIoU} & \textbf{Prec.} & \textbf{Rec.} & \textbf{mIoU} \\
    \midrule
    INSID3~\cite{cuttano2026insid3}      & \xmark & 60.74 & \textbf{82.46} & 69.19 & 61.28 & \textbf{85.20} & 69.41 \\
    RSRM~\cite{sun2026rsrm}              & \xmark & 72.48 & \secbest{77.83} & 76.69 & 73.36 & 80.30 & 77.76 \\
    HERA~\cite{ma2026hera}               & \xmark & 76.40 & 76.73 & \secbest{77.46} & 77.81 & \secbest{80.82} & \secbest{79.60} \\
    MPA~\cite{nie2026mpa}                & \xmark & 78.74 & 61.69 & 72.46 & 78.57 & 68.42 & 75.53 \\
    DefectSAM~\cite{yan2025defectsam}    & \cmark & 72.46 & 53.45 & 67.07 & 75.64 & 63.94 & 72.73 \\
    DINO-AugSeg~\cite{xu2026dinoaugseg}  & \cmark & \secbest{79.13} & 69.41 & 74.33 & \secbest{80.67} & 74.38 & 78.16 \\
    \rowcolor{oursbg}
    \textbf{SMIA (Ours)}                 & \cmark & \textbf{85.48} & 71.93 & \textbf{78.61}\,\gain{1.15} & \textbf{85.93} & 77.49 & \textbf{81.80}\,\gain{2.19} \\
    \bottomrule
  \end{tabular}%
  }}
\end{table}

Averaged across the five clinical benchmarks (\cref{tab:med_pr}), SMIA achieves the highest foreground precision at both budgets while INSID3 has the highest recall. These datasets span $7.54\%$--$25.18\%$ foreground occupancy. On BUSI breast ultrasound ($7.54\%$ mean lesion area amid acoustic speckle, with target-free normal scans making up $40\%$ of supports), 10-shot foreground precision on the full query set is $54.95\%$ for HERA versus $78.37\%$ for SMIA, which also raises recall from $31.58\%$ at 5-shot to $53.02\%$ at 10-shot. On colonoscopy polyps (CVC-ClinicDB, Kvasir-SEG) and dermoscopic skin lesions (ISIC 2018), which occupy $9.12\%$--$24.14\%$ of image area with low-contrast borders near mucosal folds or hair, SMIA improves 10-shot foreground precision over DINO-AugSeg by $+6.46$, $+5.14$, and $+8.26$~pp to $79.67\%$, $87.73\%$, and $88.09\%$. On Chest X-ray bilateral lungs ($25.18\%$ foreground area, stable anatomy), SMIA reaches $94.38\%$ mIoU at $N=5$ and $94.63\%$ at $N=10$, compared with $92.78\%$ and $93.37\%$ for DINO-AugSeg, the strongest baseline at both budgets.

\Cref{fig:supp_qual} shows five medical and industrial examples. In Rows 1--2, the pale skin lesion blends into surrounding skin and the flat polyp sits near bright endoscopic reflections. Baseline predictions spill into normal tissue or miss part or all of the target, whereas SMIA closely follows the target contours, achieving foreground IoU values of 90.05\% and 83.41\%, respectively, compared with at most $30.54\%$ for baselines. Rows 3--5 show broken capsule digits, swapped cable cores, and a thin scratch on the screw neck: baselines miss defects or flag intact structures at $0.00\%$--$23.47\%$ foreground IoU, whereas SMIA reaches 80.49\%, 84.36\%, and 86.27\% foreground IoU.

\begin{table}[!t]
  \caption{\textbf{Batch-1 inference latency and peak allocated GPU memory on one RTX 5090.} Median single-image latency (ms) and peak PyTorch allocated memory (MiB) over 80 shared queries per task across three post-warmup runs; lowest and second-lowest values are \textbf{bolded} and \secbest{underlined}.}
  \label{tab:supp_efficiency}
  \centering
  {\setlength{\aboverulesep}{0pt}\setlength{\belowrulesep}{0pt}\renewcommand{\arraystretch}{1.15}\setlength{\tabcolsep}{4.2pt}%
  \resizebox{\linewidth}{!}{%
  \begin{tabular}{lrrrr}
    \toprule
    \rowcolor{tblhead}
    & \multicolumn{2}{c}{\textbf{CVC-ClinicDB}} & \multicolumn{2}{c}{\textbf{MVTec Bottle}} \\
    \cmidrule(lr){2-3} \cmidrule(lr){4-5}
    \rowcolor{tblhead}
    \textbf{Method} & \textbf{Lat.\ (ms)} & \textbf{Mem.\ (MiB)} & \textbf{Lat.\ (ms)} & \textbf{Mem.\ (MiB)} \\
    \midrule
    INSID3~\cite{cuttano2026insid3}      & 277.57 & 3,283.30 &  73.28 & 2,673.37 \\
    RSRM~\cite{sun2026rsrm}              & 509.43 &   926.43 & 299.96 &   762.33 \\
    HERA~\cite{ma2026hera}               & 249.44 & 1,603.78 & 158.85 & 1,537.80 \\
    MPA~\cite{nie2026mpa}                &  53.56 & \secbest{407.73} &  60.09 & \secbest{406.44} \\
    DefectSAM~\cite{yan2025defectsam}    &  74.52 & 2,819.17 &  83.39 & 2,819.17 \\
    DINO-AugSeg~\cite{xu2026dinoaugseg}  & \secbest{22.70} & 1,475.48 & \secbest{35.95} & 1,475.48 \\
    \rowcolor{oursbg}
    \textbf{SMIA (Ours)}                 & \textbf{13.66} & \textbf{165.93} & \textbf{23.50} & \textbf{219.85} \\
    \bottomrule
  \end{tabular}%
  }}
\end{table}

\begin{table*}[!t]
  \caption{\textbf{One-shot evaluation and cross-budget scaling across the 20-task benchmark} (\%, differences computed before rounding). \textbf{(a)} Full-query 1-shot ($N=1$) segmentation mIoU and six-dataset foreground precision (\emph{Prec.}), reported as mean $\pm$ SD. \textbf{(b)} Mean mIoU across $N \in \{1, 5, 10\}$ on the common query set outside the 10-shot supports, showing five-task \emph{Medical}, 15-class \emph{MVTec AD}, and six-dataset \emph{Overall} summaries alongside signed Overall changes $\Delta_{1\to 5}$ and $\Delta_{1\to 10}$. \emph{Sup.-Free} indicates query-only inference; best and second-best scores in each metric column are \textbf{bolded} and \secbest{underlined}.}
  \label{tab:supp_oneshot}
  \vspace{-5pt}
  \centering
  {\renewcommand{\std}[1]{{\small\textcolor{stdgray}{\,$\pm$\,#1}}}\renewcommand{\bstd}[1]{{\small\textcolor{stdgray}{\,$\pm$\,#1}}}%
  \setlength{\aboverulesep}{0pt}\setlength{\belowrulesep}{0pt}\renewcommand{\arraystretch}{1.06}\setlength{\tabcolsep}{2.9pt}%
  \resizebox{\textwidth}{!}{%
  \begin{tabular}{lcrrrrrrrr}
    \toprule
    \rowcolor{tblsub}
    \multicolumn{10}{l}{\textit{\textbf{(a) 1-Shot Benchmark on Full Query Sets ($N = 1$)}}} \\
    \midrule
    \rowcolor{tblhead}
    & & \multicolumn{5}{c}{\textbf{Medical Imaging Benchmarks (5 Tasks)}} & \textbf{Industrial (15)} & \multicolumn{2}{c}{\textbf{6-Dataset Summary}} \\
    \cmidrule(lr){3-7} \cmidrule(lr){8-8} \cmidrule(lr){9-10}
    \rowcolor{tblhead}
    \textbf{Method} & \textbf{Sup.-Free} & \multicolumn{1}{c}{\textbf{BUSI}} & \multicolumn{1}{c}{\textbf{Kvasir-SEG}} & \multicolumn{1}{c}{\textbf{CVC-ClinicDB}} & \multicolumn{1}{c}{\textbf{ISIC 2018}} & \multicolumn{1}{c}{\textbf{Chest X-ray}} & \multicolumn{1}{c}{\textbf{MVTec AD}} & \multicolumn{1}{c}{\textbf{Prec.}} & \multicolumn{1}{c}{\textbf{Overall mIoU}} \\
    \midrule
    INSID3~\cite{cuttano2026insid3}      & \xmark & 41.08\std{12.80} & \secbest{78.02}\std{4.29} & 55.99\std{9.11}  & 65.22\std{15.31} & 85.35\std{0.27} & 50.98\std{1.47} & 48.03\std{5.57} & 62.77\std{4.63} \\
    RSRM~\cite{sun2026rsrm}              & \xmark & \secbest{54.46}\std{3.87}  & \textbf{78.10}\bstd{4.27} & 65.95\std{12.35} & \textbf{73.01}\bstd{3.42}  & 91.13\std{0.92} & 52.33\std{0.79} & 57.89\std{5.82} & \textbf{69.16}\bstd{3.15} \\
    HERA~\cite{ma2026hera}               & \xmark & \textbf{58.90}\bstd{4.71}  & 72.89\std{7.01}  & \textbf{66.82}\bstd{7.59} & \secbest{71.96}\std{5.50}  & 84.09\std{7.90} & 55.18\std{0.88} & 61.63\std{7.33} & \secbest{68.31}\std{3.44} \\
    MPA~\cite{nie2026mpa}                & \xmark & 48.61\std{1.21}  & 57.11\std{8.63}  & 59.90\std{8.35}  & 47.16\std{5.30}  & \secbest{91.37}\std{0.58} & \secbest{58.68}\std{0.56} & 65.10\std{5.88} & 60.47\std{1.71} \\
    DefectSAM~\cite{yan2025defectsam}    & \cmark & 46.51\std{0.59}  & 46.34\std{4.19}  & 45.60\std{3.12}  & 33.91\std{18.33} & 69.58\std{7.14} & 51.89\std{0.75} & 44.10\std{3.84} & 48.97\std{2.04} \\
    DINO-AugSeg~\cite{xu2026dinoaugseg}  & \cmark & 47.66\std{1.77}  & 66.28\std{12.37} & \secbest{66.44}\std{6.27} & 52.96\std{18.24} & 86.67\std{4.00} & 56.32\std{1.19} & \secbest{69.75}\std{5.13} & 62.72\std{3.68} \\
    \rowcolor{oursbg}
    \textbf{SMIA (Ours)}                 & \cmark & 52.59\std{5.78}  & 67.33\std{11.30} & 64.99\std{12.64} & 60.40\std{14.12} & \textbf{93.83}\bstd{0.30} & \textbf{61.78}\bstd{1.33} & \textbf{71.89}\bstd{7.74} & 66.82\std{3.59} \\
    \bottomrule
  \end{tabular}%
  }}\\[2pt]
  {\setlength{\aboverulesep}{0pt}\setlength{\belowrulesep}{0pt}\renewcommand{\arraystretch}{1.06}\setlength{\tabcolsep}{5.2pt}%
  \resizebox{\textwidth}{!}{%
  \begin{tabular}{lcrrrrrrrrrrr}
    \toprule
    \rowcolor{tblsub}
    \multicolumn{13}{l}{\textit{\textbf{(b) Cross-Budget Mean mIoU on Common Query Sets ($N \in \{1, 5, 10\}$)}}} \\
    \midrule
    \rowcolor{tblhead}
    & & \multicolumn{3}{c}{\textbf{Medical (5 Tasks)}} & \multicolumn{3}{c}{\textbf{MVTec AD (15)}} & \multicolumn{5}{c}{\textbf{6-Dataset Overall mIoU}} \\
    \cmidrule(lr){3-5} \cmidrule(lr){6-8} \cmidrule(lr){9-13}
    \rowcolor{tblhead}
    \textbf{Method} & \textbf{Sup.-Free} & \textbf{1-Shot} & \textbf{5-Shot} & \textbf{10-Shot} & \textbf{1-Shot} & \textbf{5-Shot} & \textbf{10-Shot} & \textbf{1-Shot} & \textbf{5-Shot} & \textbf{10-Shot} & \multicolumn{1}{c}{$\boldsymbol{\Delta_{1\to 5}}$} & \multicolumn{1}{c}{$\boldsymbol{\Delta_{1\to 10}}$} \\
    \midrule
    INSID3~\cite{cuttano2026insid3}      & \xmark & 65.13 & 69.17 & 69.41 & 50.93 & 56.70 & 57.98 & 62.77 & 67.09 & 67.51 & $+4.33$  & $+4.74$ \\
    RSRM~\cite{sun2026rsrm}              & \xmark & \textbf{72.54} & 76.70 & 77.76 & 52.20 & 52.52 & 52.40 & \textbf{69.15} & 72.67 & 73.54 & $+3.52$  & $+4.38$ \\
    HERA~\cite{ma2026hera}               & \xmark & \secbest{70.95} & \secbest{77.46} & \secbest{79.60} & 55.08 & 55.34 & 58.30 & \secbest{68.31} & \secbest{73.78} & 76.05 & $+5.47$  & $+7.75$ \\
    MPA~\cite{nie2026mpa}                & \xmark & 60.83 & 72.47 & 75.53 & \secbest{58.74} & 65.12 & 67.62 & 60.48 & 71.25 & 74.21 & $+10.77$ & $+13.73$ \\
    DefectSAM~\cite{yan2025defectsam}    & \cmark & 48.40 & 67.07 & 72.73 & 51.92 & 62.20 & 68.30 & 48.98 & 66.26 & 71.99 & $+17.27$ & $+23.01$ \\
    DINO-AugSeg~\cite{xu2026dinoaugseg}  & \cmark & 64.03 & 74.33 & 78.16 & 56.42 & \secbest{66.47} & \secbest{71.58} & 62.76 & 73.02 & \secbest{77.06} & $+10.26$ & $+14.30$ \\
    \rowcolor{oursbg}
    \textbf{SMIA (Ours)}                 & \cmark & 67.83 & \textbf{78.62} & \textbf{81.80} & \textbf{61.85} & \textbf{71.45} & \textbf{76.96} & 66.84 & \textbf{77.43} & \textbf{80.99} & $+10.59$ & $+14.15$ \\
    \bottomrule
  \end{tabular}%
  }}
  \vspace{-7pt}
\end{table*}

\subsection{Depth and Subspace Width Analysis}
\label{sec:supp_depth_width}

Decomposing the depth sweep in Main Paper Tab.~4 into precision and recall shows how adapting earlier blocks changes segmentation errors. On the six-dataset summary at $N=10$, adapting the last 4 blocks ($170.9\mathrm{K}$ parameters), last 8 blocks, or all 12 blocks yields nearly constant foreground recall ($74.79\%$, $74.94\%$, and $74.80\%$), while foreground precision rises steadily from $83.62\%$ to $85.12\%$ and $85.54\%$ ($80.89\% \to 84.16\% \to 84.48\%$ across the ten MVTec AD object categories, and $84.22\% \to 85.93\%$ across the five medical datasets).

Precision and recall also remain stable across the subspace-width and off-diagonal mixing variants in Main Paper Tab.~3. Even at the smallest width $r=4$ ($243.5\mathrm{K}$ parameters, less than half of the ${\approx}0.51\mathrm{M}$-parameter PEFT baselines), and across all four SMIA configurations spanning $r \in \{4, 8, 16\}$ ($243.5\mathrm{K}$--$1.24\mathrm{M}$ parameters), three-dataset precision stays within $88.41\%$--$88.77\%$ and recall within $68.71\%$--$69.19\%$, outperforming AdaptFormer ($84.51\%$ precision, $67.59\%$ recall), LoRA ($87.03\%$, $66.99\%$), and ParaX ($86.22\%$, $68.03\%$). On the ten MVTec object classes, SMIA reaches $83.80\%$--$86.01\%$ precision ($85.05\%$ at $r=8$) compared with $79.28\%$--$82.88\%$ for the PEFT baselines. Adding off-diagonal mixing $\mathbf{B}_h$ at $r=8$ introduces $32{,}256$ parameters across 12 blocks and lifts three-dataset precision from $88.41\%$ to $88.77\%$. Doubling the width to $r=16$ at $1.24\mathrm{M}$ parameters raises mIoU from $76.74\%$ to $77.29\%$ on the ten MVTec object classes and from $80.90\%$ to $81.96\%$ on CVC-ClinicDB, while ISIC 2018 mIoU drops from $79.33\%$ to $77.98\%$, leaving the three-dataset mean nearly unchanged as in Main Paper Tab.~3.

\subsection{Inference Latency and Memory Footprint}
\label{sec:supp_efficiency}

\Cref{tab:supp_efficiency} benchmarks single-image inference on CVC-ClinicDB ($384{\times}288$ native, $512^2$ SMIA canvas) and MVTec Bottle ($900^2$ native, $768^2$ SMIA canvas), with baselines running at their official resolutions from \cref{sec:supp_baselines} under shared 10-shot supports where applicable. We measure median wall-clock latency from a decoded RGB array to the native-resolution binary mask, together with peak PyTorch-allocated GPU memory. Timing covers preprocessing, the forward pass, and mask restoration, excluding disk I/O, model loading, and one-time support setup. Matchers INSID3, RSRM, and HERA require $73.28$--$509.43$~ms and up to $3{,}283.30$~MiB due to ViT-B/L feature extraction and cross-image matching, while multi-scale decoders DefectSAM and DINO-AugSeg allocate $2{,}819.17$ and $1{,}475.48$~MiB. With a single support-free pass through ViT-S+/16 and its 385-parameter linear head, SMIA achieves both the lowest latency and smallest memory footprint at 13.66~ms and 165.93~MiB on CVC-ClinicDB and 23.50~ms and 219.85~MiB on MVTec Bottle, running $1.53\times$--$1.66\times$ faster than DINO-AugSeg at $6.7\times$--$8.9\times$ lower memory.

\subsection{One-Shot Evaluation and Support-Budget Scaling}
\label{sec:supp_oneshot}

SMIA fits shared in-block conditional weights from support masks (Main Paper Sec.~3.3) rather than retaining support exemplars for query-time matching. To test how well SMIA segments unseen images when each task provides only a single annotated foreground support ($N=1$), we keep all 12 adapted blocks, width $r=8$, and 2,000 joint updates unchanged.

\Cref{tab:supp_oneshot}a reports the 1-shot benchmark. On six-dataset Overall mIoU, SMIA reaches $66.82\%$, trailing matchers RSRM ($69.16\%$) and HERA ($68.31\%$) while leading all support-free methods: using $1.7\%$ as many trainable parameters as DINO-AugSeg, it exceeds DINO-AugSeg ($62.72\%$) by $+4.10$~pp and DefectSAM ($48.97\%$) by $+17.85$~pp, and also outperforms MPA ($60.47\%$) and INSID3 ($62.77\%$). Across all seven methods, SMIA attains the highest foreground precision on the six-dataset summary ($71.89\%$), across the five medical datasets ($73.10\%$), and on MVTec AD ($65.86\%$). On MVTec AD, SMIA leads all compared methods at $61.78\%$ mIoU ($+3.10$~pp over MPA; $24.63\%$ vs.\ $18.43\%$ foreground IoU) and ranks first on both textures ($69.09\%$) and object categories ($58.12\%$). On Chest X-ray, it reaches $93.83\%$ mIoU ($+2.46$~pp over MPA). Across the 20 tasks, SMIA ranks first on 11 among all seven methods and on 18 among support-free methods; regular surface patterns or stable anatomy may help explain its strong 1-shot results on MVTec AD and Chest X-ray. On the four lesion and polyp benchmarks, however, training from a single scan is more sensitive to target and background variation: SMIA shows high precision with conservative recall on BUSI and Kvasir-SEG ($77.09\%$ and $82.21\%$ precision vs.\ $12.63\%$ and $51.75\%$ recall), higher recall with lower precision across unseen backgrounds on CVC-ClinicDB and ISIC 2018 ($79.49\%$ and $86.82\%$ recall vs.\ $52.20\%$ and $58.24\%$ precision), and $11.30\%$--$14.12\%$ mIoU SD across Kvasir-SEG, CVC-ClinicDB, and ISIC 2018 depending on the selected support image. Direct query--support matching on ViT-B/L backbones gives RSRM and HERA higher 1-shot five-dataset medical mIoU ($72.53\%$ and $70.93\%$ vs.\ $67.83\%$ for SMIA).

Because full-inventory evaluation at $N=1$ and $N=5$ includes scans that enter the nested 10-shot support set (Main Paper Sec.~4.1), \cref{tab:supp_oneshot}b re-evaluates $N \in \{1, 5, 10\}$ on the identical query set outside the 10-shot supports. Other parameter-adapted baselines also improve substantially from $N=1$ to $N=5$ ($+10.26$~pp for DINO-AugSeg, $+10.77$~pp for MPA, and $+17.27$~pp for DefectSAM) from lower 1-shot starting points; SMIA starts $+4.07$~pp above DINO-AugSeg at $N=1$ ($66.84\%$ vs.\ $62.76\%$), gains $+10.59$~pp at $N=5$ ($77.43\%$) and $+14.15$~pp at $N=10$ ($80.99\%$), and stays $+3.93$ to $+4.40$~pp ahead of DINO-AugSeg at all three budgets. While matchers RSRM ($69.15\%$) and HERA ($68.31\%$) lead at $N=1$, they gain only $+3.52$ and $+5.47$~pp overall from $N=1$ to $N=5$ and stay nearly flat on MVTec AD ($+0.32$ and $+0.26$~pp), likely because recurring intact structures remain hard to distinguish from defects using positive references alone. As a result, SMIA overtakes both matchers by $N=5$ ($77.43\%$ vs.\ $73.78\%$ for HERA and $72.67\%$ for RSRM) and widens the lead at $N=10$ ($80.99\%$ vs.\ $76.05\%$ and $73.54\%$). With its $0.51\mathrm{M}$ parameters and 2,000-step schedule fixed, broader target coverage and added normal supports on BUSI and MVTec AD raise both Overall precision ($71.85\% \to 84.90\% \to 85.54\%$) and recall ($59.25\% \to 68.01\% \to 74.80\%$). Improvements from $N=1$ to $N=5$ are largest on ISIC 2018 ($+19.59$~pp), Kvasir-SEG ($+12.95$~pp), CVC-ClinicDB ($+11.34$~pp), MVTec AD ($+9.60$~pp), and BUSI ($+9.53$~pp), whereas Chest X-ray is already near saturation at $N=1$ ($93.83\% \to 94.38\% \to 94.63\%$).

\section{Representation and Mechanistic Evidence}
\label{sec:supp_mechanism}

\subsection{Frozen Representation Diagnostics}
\label{sec:supp_diag}

We inspect patch representations at the final LayerNorm of frozen DINOv3 ViT-S+/16 near subtle boundaries. On MVTec Cable ($768\times 768$ canvas), we form two cross-image comparisons across four test scans, each pairing a $2\times 2$-patch damaged inner-insulation region with $2\times 2$-patch intact inner-insulation and outer-sheath cut regions in another scan. In both comparisons, the frozen features of damaged inner insulation share higher median final-layer cosine similarity with those of intact inner insulation than with those of outer-sheath cuts (Group~1: $0.707$ vs.\ $0.552$; Group~2: $0.668$ vs.\ $0.553$), tracking component identity more closely than defect status.

We also test a five-support $\lambda=10^{-3}$ ridge-regularized balanced BCE probe on two benign BUSI queries ($32\times 32$ grid): in one query, a lesion-containing patch and vertically adjacent stroma patch have final-layer cosine $0.962$ and logits $+2.848$ and $-1.880$; in the other query with different speckle statistics, the boundary pair has cosine $0.958$ and $48.4\%$ foreground coverage in the lesion-side patch, yet both logits are negative ($-6.166$ and $-8.052$) and the readout misses the margin.

Within individual attention heads, aggregated messages can already show local target--background differences before output projection. Main Paper Fig.~1 shows Block~9 Head~2 for BUSI and Block~9 Head~5 for Capsule. For the 64-dimensional message $\mathbf{m}_t$ at position~$t$, we define $d_t$ as its mean cosine similarity to up to eight pure foreground references $F$ minus its mean cosine similarity to up to eight pure background references $B$ within the expanded foreground bounding box:
\begin{equation}
  d_t = \frac{1}{|F|}\sum_{i\in F}\cos(\mathbf{m}_t, \mathbf{m}_i) - \frac{1}{|B|}\sum_{j\in B}\cos(\mathbf{m}_t, \mathbf{m}_j).
  \label{eq:supp_contrast}
\end{equation}
For the diagnostics in Main Paper Fig.~1 and \cref{fig:supp_motivation}, we compare frozen DINOv3, rank-8 Q/V LoRA across all 12 blocks ($147.8\mathrm{K}$ trainable parameters), and width-8 SMIA on Blocks~9--12 ($170.9\mathrm{K}$ trainable parameters, with Blocks~1--8 unchanged), fitting linear heads on the same 10 supports per case. In the six examples in \cref{fig:supp_motivation}, selected frozen attention heads show target--background contrast even where frozen final-layer similarity extends into neighboring tissue or texture. LoRA misses most of the foreground in the BUSI, CVC-ClinicDB, Kvasir-SEG, ISIC 2018, and Carpet examples and leaves a gap within one lung field on Chest X-ray, whereas SMIA follows the annotated targets more closely. On an additional CVC-ClinicDB query, frozen Block-11 Head-1 contrast averages $+1.06$ on non-reference polyp patches vs.\ $-0.43$ on nearby mucosa, and frozen, LoRA, and SMIA readouts reach $35.5\%$, $5.2\%$, and $91.1\%$ foreground recall, respectively.

\begin{figure*}[!t]
  \centering
  \includegraphics[width=0.96\textwidth]{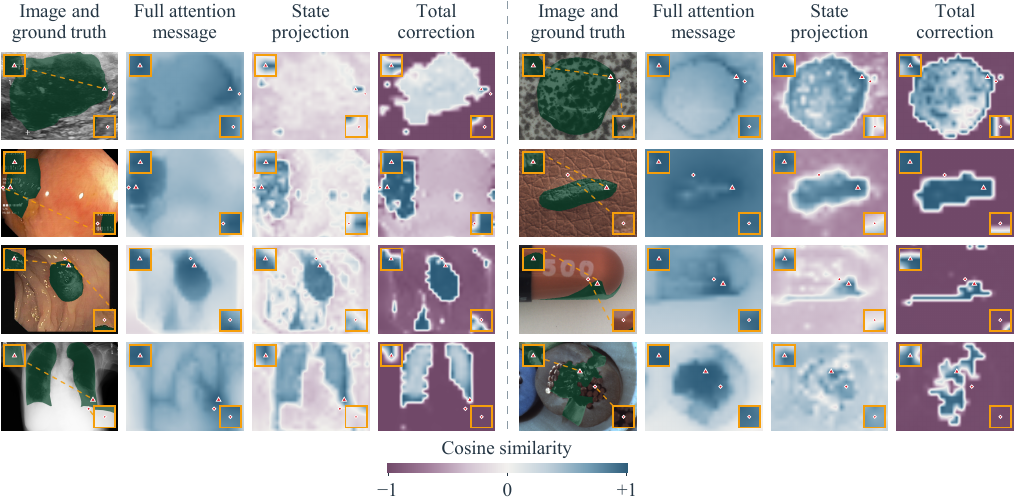}
  \vspace{-4pt}
  \caption{\textbf{Directions of attention messages, state projections, and corrections across eight tasks at Block~12 ($N=10$).} Left, top to bottom: BUSI, Kvasir-SEG, CVC-ClinicDB, and Chest X-ray. Right, top to bottom: MVTec Tile, Leather, Capsule, and Cable. Green overlays mark ground truth, and amber insets magnify nearby foreground ($A$, red triangle) and background ($B$, red circle) boundary patches; heatmaps plot cosine similarity to $A$ in $[-1, +1]$ for the full six-head attention message, state projection, and total correction (\cref{sec:supp_forward}).}
  \label{fig:supp_dir}
  \vspace{-7pt}
\end{figure*}

\subsection{Counterfactual State-Swap Protocol}
\label{sec:supp_swap}

Using our 12-block $N=10$ SMIA checkpoints, donor matching operates on concatenated six-head Block-12 messages prior to $\mathbf{M}_h$ over unpadded pure foreground ($100\%$ mask coverage) and pure background ($0\%$ coverage) patches. Within candidate donor queries of the same task, we retain patches with six-head message cosine distance $d \le 0.20$ ($d \le 0.30$ on CVC-ClinicDB, Kvasir-SEG, and Chest X-ray) and donor-to-target message $\ell_2$-norm ratio $r_{\mathrm{msg}} \in [0.8, 1.25]$. The same-label donor $s$ and opposite-label donor $o$ come from the same donor scan and satisfy $|d_s - d_o| \le 0.05$ and $|\log r_{\mathrm{msg},s} - \log r_{\mathrm{msg},o}| \le \log(1.1)$; among feasible shortlisted pairs, we select the one minimizing $\max(d_s, d_o)$ across donor scans. With the post-attention Block-12 feature and six-head messages fixed at the target patch, we replace all six state projections $\mathbf{a}_{t1}, \dots, \mathbf{a}_{tH}$ with those from the donor and denote the unrescaled correction by $\boldsymbol{\delta}'$. To separate direction from magnitude, we add either the norm-matched correction $\boldsymbol{\delta}' \cdot (\|\boldsymbol{\delta}\|_2 / \|\boldsymbol{\delta}'\|_2)$ or the magnitude-only control $\boldsymbol{\delta} \cdot (\|\boldsymbol{\delta}'\|_2 / \|\boldsymbol{\delta}\|_2)$ to the fixed post-attention feature, then pass the result through Block~12's frozen FFN, final LayerNorm, and linear head.

Probability maps in Main Paper Fig.~5 and \cref{fig:supp_swap} combine independent swaps at matched patches and keep original predictions elsewhere. \Cref{fig:supp_swap} shows eight additional target patches (four foreground, four background). For both same-label and opposite-label state swaps, we rescale the recomputed correction to the norm of the target patch's original correction. At all eight marked patches, same-label swaps preserve the predicted class, while opposite-label swaps change the correction direction and flip the predicted class.

\subsection{Extended Forward-Pass Direction Cases}
\label{sec:supp_forward}

Complementing Main Paper Fig.~4, \cref{fig:supp_dir} reports Block-12 cosine alignment across nearby foreground and background boundary pairs $(A, B)$ on four medical tasks and four industrial categories. Across all eight pairs, concatenated six-head messages remain strongly aligned with $\cos(\mathbf{m}_A, \mathbf{m}_B) \in [0.766, 0.944]$, local state projections span $\cos(\mathbf{a}_A, \mathbf{a}_B) \in [-0.184, 0.535]$, and SMIA maps them to opposing updates with $\cos(\boldsymbol{\delta}_A, \boldsymbol{\delta}_B) \in [-0.997, -0.645]$. Opposing updates do \emph{not} require $\mathbf{a}_A$ and $\mathbf{a}_B$ to be negatively aligned: on CVC-ClinicDB and MVTec Cable inner-insulation cut, message cosines of $0.766$ and $0.895$ pair with positively aligned state projections of $+0.216$ and $+0.535$, yet the corresponding correction vectors point in opposing directions, with cosine similarities of $-0.885$ and $-0.972$. Under SMIA's bilinear parameterization, even for fixed messages, the induced linear map from stacked state projections to $\boldsymbol{\delta}_t$ need not preserve angles, so positively aligned state projections can still yield opposing updates.